\documentclass{article}

\PassOptionsToPackage{square,sort,comma,numbers}{natbib}

\usepackage[preprint,nonatbib]{openrobotlab}

\usepackage[utf8]{inputenc} % allow utf-8 input
\usepackage[T1]{fontenc}    % use 8-bit T1 fonts
\usepackage{hyperref}       % hyperlinks
\usepackage{url}            % simple URL typesetting
\usepackage{booktabs}       % professional-quality tables
\usepackage{amsfonts}       % blackboard math symbols
\usepackage{nicefrac}       % compact symbols for 1/2, etc.
\usepackage{microtype}      % microtypography
\usepackage{xcolor}         % colors
\usepackage{graphicx}
\usepackage{multirow}
\usepackage{caption}
\usepackage{subcaption}
\usepackage{wrapfig}
\usepackage{array}
\usepackage{enumitem}
\usepackage{algorithm}
\usepackage{fancyvrb}
\usepackage{fvextra}
\usepackage[frozencache,cachedir=.]{minted}
\usepackage[most,breakable]{tcolorbox}
\usepackage{bm}
\usepackage{longtable}

\definecolor{mygray}{gray}{0.9}
\definecolor{argument}{RGB}{51,188,255}
\definecolor{function}{RGB}{255,166,77}

\newcommand{\funcc}[1]{{\color{function}#1}}
\newcommand{\argc}[1]{{\color{argument}#1}}

\newcommand{\red}[1]{{\color{red}#1}}

\newcommand{\CheckMark}{\includegraphics[scale=0.06]{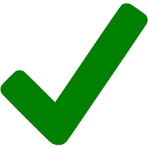}}
\newcommand{\CrossMark}{\includegraphics[scale=0.06]{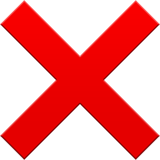}}
\newcommand{\ie}{\textit{i.e.,} }

\newcolumntype{P}[1]{>{\centering\arraybackslash}p{#1}} % See https://tex.stackexchange.com/questions/157389/how-to-center-column-values-in-a-table



\begin{document}
\title{GRUtopia: Dream General Robots in a City at Scale}

% The \author macro works with any number of authors. There are two commands
% used to separate the names and addresses of multiple authors: \And and \AND.
%
% Using \And between authors leaves it to LaTeX to determine where to break the
% lines. Using \AND forces a line break at that point. So, if LaTeX puts 3 of 4
% authors names on the first line, and the last on the second line, try using
% \AND instead of \And before the third author name.

\author{%
Hanqing Wang$^{1*}$, Jiahe Chen$^{1,2*}$, Wensi Huang$^{1,3*}$, Qingwei Ben$^{1,4*}$, Tai Wang$^{1*}$, Boyu Mi$^{1,2*}$\\
\textbf{Tao Huang$^{1}$, Siheng Zhao$^{1,5}$, Yilun Chen$^{1}$, Sizhe Yang$^{1,6}$, Peizhou Cao$^{1,7}$, Wenye Yu$^{1,3}$}\\
\textbf{Zichao Ye$^{1}$, Jialun Li$^{1}$, Junfeng Long$^{1}$, Zirui Wang$^{1,2}$, Huiling Wang$^{1}$, Ying Zhao$^{1}$}\\
\textbf{Zhongying Tu$^{1}$, Yu Qiao$^{1}$, Dahua Lin$^{1,6}$, Jiangmiao Pang$^{1\dagger}$} 
\vspace{+2mm} \\
$^1$OpenRobotLab, Shanghai AI Laboratory \\
$^2$Zhejiang University,
$^3$Shanghai Jiao Tong University,
$^4$Tsinghua University \\
$^5$Nanjing University, 
$^6$The Chinese University of Hong Kong, 
$^7$Xidian University 
\vspace{0.2ex} \\
$^\ast$equal contribution\quad
$^\dagger$corresponding author 
}
\maketitle
\vspace{-6ex}
\begin{figure}[h]
\begin{center}
% \begin{overpic} 
% [width=\linewidth]
% {example-image-a}
% \end{overpic}
\includegraphics[width=0.95\linewidth]{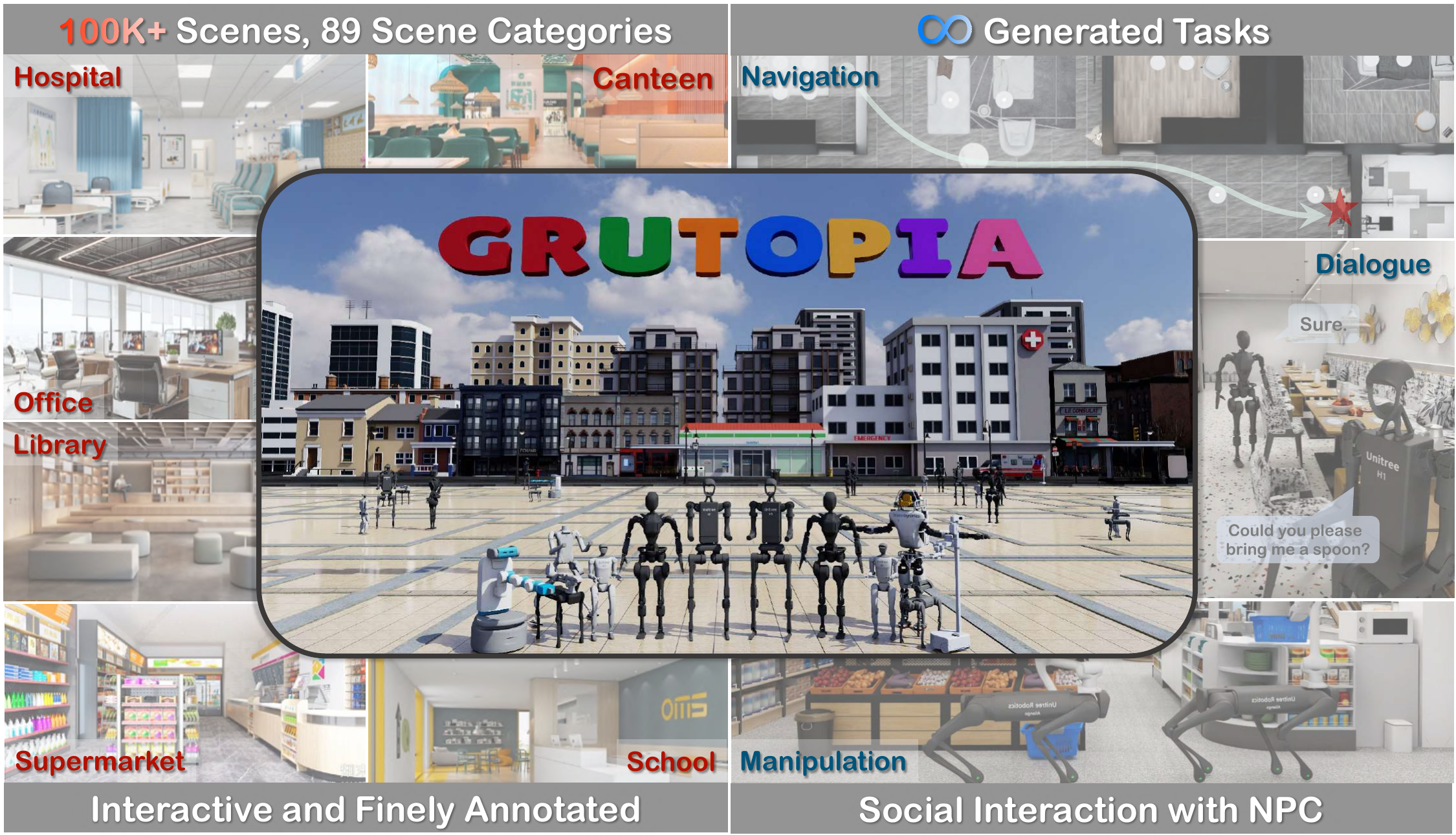}
\end{center}
\vspace{-0.08in}
\caption{
Key features of GRUtopia.
}
\vspace{-2ex}
\label{fig:example}
\end{figure}
\begin{abstract}
\vspace{-5pt}
Recent works have been exploring the scaling laws in the field of Embodied AI. Given the prohibitive costs of collecting real-world data, we believe the Simulation-to-Real (Sim2Real) paradigm is a crucial step for scaling the learning of embodied models.
This paper introduces project \textit{GRUtopia}, the first simulated interactive 3D society designed for various robots.
It features several advancements:
(a) The scene dataset, \textit{GRScenes}, includes 100k interactive, finely annotated scenes, which can be freely combined into city-scale environments. In contrast to previous works mainly focusing on home, GRScenes covers 89 diverse scene categories, bridging the gap of service-oriented environments where general robots would be initially deployed.
(b) \textit{GRResidents}, a Large Language Model (LLM) driven Non-Player Character (NPC) system that is responsible for social interaction, task generation, and task assignment, thus simulating social scenarios for embodied AI applications.
(c) The benchmark, \textit{GRBench}, supports various robots but focuses on legged robots as primary agents and poses \textit{moderately challenging} tasks involving Object Loco-Navigation, Social Loco-Navigation, and Loco-Manipulation.
We hope that this work can alleviate the scarcity of high-quality data in this field and provide a more comprehensive assessment of Embodied AI research. 
The project is available at~\url{https://github.com/OpenRobotLab/GRUtopia}.

\end{abstract}

\vspace{-2ex}
\section{Introduction}
\label{sec:intro}
\vspace{-1.5ex}

Scaling law has seen significant success in NLP and CV, inspiring the robotics community to explore its form in robot learning.
A straightforward approach is collecting real-world robot action trajectories, such as recent efforts in Open X-Embodiment~\cite{padalkar2023open} and DROID~\cite{khazatsky2024droid}.
However, such attempts pose persistent challenges regarding the data collection costs and generalization problems across different hardware platforms.
We believe simulation is a crucial step in addressing these issues.
Previous works have demonstrated the possibility of learning specific low-level policies, such as Legged Gym~\cite{agileleggedrobot,walkinminutes}, ManiSkill~\cite{gu2023maniskill2}, and Orbit~\cite{mittal2023orbit}. Many works have explored Embodied AI in simulation~\cite{behavior, arnold, alfred, habitat2020sim2real}.
However, existing platforms exhibit limited diversity and complexity regarding two critical aspects: scenarios and tasks, thereby struggling to meet the demand for achieving policy generalization.

To address the above limitations, this paper presents GRUtopia, \textbf{the first simulated interactive 3D society} designed for various robots that serve humans. GRUtopia distinguishes itself from previous platforms in three key aspects: (a) GRScenes, a large-scale scene dataset capable of constructing city-scale landscapes. It includes 100K fully interactive, finely annotated scenes covering 89 functional categories\footnote{Due to license issues, we will initially release 100 scenes across 7 building types. We are actively working on cleaning the remaining data and plan to release it progressively over time.}; (b) GRResidents, an NPC system leveraging LLMs to generate diverse social characters for interaction, task creation, and assignment; and (c) GRBench, a benchmark featuring “moderately challenging” tasks aligned with current algorithm capabilities.

Firstly, GRScenes significantly expands the scope of environments in which robots can operate. Previous works have predominantly focused on developing general agents for home environments~\cite{puig2018virtualhome, homerobot, alfred}. Beyond housework, we aim to extend the capabilities of general robots to service-oriented scenarios such as supermarkets and hospitals, where they can be initially deployed. GRScenes encompasses both indoor and outdoor environments, including restaurants, supermarkets, offices, libraries, museums, hospitals, exhibition halls, amusement parks, homes, \emph{etc.} These scenes feature physically realistic materials, detailed external appearances and accessible internal structures, complete with furniture arrangements. The dataset includes numerous high-quality, part-level modeled objects, ensuring that the scenes are fully dynamic and interactive. We provide fine-grained, hierarchical, multi-modal annotations for both scenes and objects, covering levels from the overall scene, indoor regions, objects, down to individual parts.

Secondly, GRResidents, our NPC system, introduces a new dimension to human-robot interaction within simulations. The inclusion of NPCs is motivated by the goal that robots are ultimately meant to serve humans, and interaction with users is often helpful or necessary for task completion, such as resolving ambiguities according to user preferences. GRResidents integrates an LLM agent framework with a hierarchical scene perception module. It possesses comprehensive knowledge about the environment, including attributes, appearance, and structural information of objects provided in our dataset. NPCs can infer spatial relationships between objects, understand scene semantics, observe the activities of other agents in real-time, and engage in dynamic dialogues and task assignments based on this information. Resorting to the powerful NPCs, GRUtopia can generate an infinite number of scene-aware embodied tasks.

Lastly, GRBench serves as a comprehensive evaluation tool for assessing robot agents’ capabilities. To benchmark the robot agents’ ability to handle daily tasks, GRBench comprises three benchmarks: Object Loco-Navigation, Social Loco-Navigation, and Loco-Manipulation. These benchmarks are designed to progressively increase in difficulty, demanding enhanced robotic skills. We prioritize legged robots as primary agents due to their superior cross-terrain capability. However, in large-scale scenarios, it is challenging for current algorithms to simultaneously perform high-level perception, planning, and low-level control while achieving satisfactory results. Inspired by recent progress, which demonstrates the feasibility of training highly accurate policies for individual skills in simulation, the initial version of GRBench focuses on high-level tasks and officially provides learning-based control policies as APIs, such as walking and pick-and-place. Consequently, our benchmark offers a more physically genuine setting that narrows the gap between simulation and the real world.

We conduct extensive experiments to analyze the performance of our NPCs and control APIs, proposing LLM and VLM-driven baselines to validate the soundness of our benchmark design and study the ability of existing LLM or VLM-based agents to tackle embodied tasks. The experimental results reveal that integrating realistic actions with high-level planning increases task difficulty, presenting a key challenge for previous embodied algorithms when applied to real-world scenarios.
It also shows that our constructed benchmark and evaluation metrics exhibit well-defined difficulty granularity and are capable of meeting the research needs across various fields and hierarchies. We hope this platform, with our continuous efforts to scale up the diversity of scenes and tasks on top, can benefit the community.

\vspace{-2ex}
\section{Related Work}
\vspace{-1.5ex}
\label{sec:relatedwork}
\begin{table}[t]

\centering
\renewcommand\arraystretch{1.1}
\caption{\small
Comparison of GRUtopia with other platforms in terms of Scene, Object, Platform, and Benchmark.
} % \caption
% \vspace{1pt}
\resizebox{\linewidth}{!}{ %< auto-adjusts font size to fill line
\begin{tabular}{@{}ccP{0.05\textwidth}P{0.05\textwidth}P{0.05\textwidth}P{0.05\textwidth}P{0.05\textwidth}P{0.05\textwidth}P{0.05\textwidth}P{0.05\textwidth}P{0.05\textwidth}P{0.05\textwidth}P{0.05\textwidth}P{0.05\textwidth}P{0.05\textwidth}|@{}}

~ & ~ & \rotatebox{30}{\textbf{GRUtopia}} & \rotatebox{30}{Behavior-1K~\cite{behavior}}  & \rotatebox{30}{Arnold~\cite{arnold}} & \rotatebox{30}{ALFRED~\cite{alfred}} & \rotatebox{30}{Vis. Room Rearr.~\cite{visualroomrearrangement}} & \rotatebox{30}{ManipulaTHOR~\cite{manipulathor}} & \rotatebox{30}{ProcTHOR-10k~\cite{procthor}} & \rotatebox{30}{VLN-CE~\cite{vlnce}} & \rotatebox{30}{HomeRobot~\cite{homerobot}} & \rotatebox{30}{Social Navigation~\cite{socialnavigation}} & \rotatebox{30}{Maniskill2~\cite{gu2023maniskill2}} & \rotatebox{30}{VLN~\cite{room2room}} & \multicolumn{1}{c}{\rotatebox{30}{CVDN~\cite{cvdn}}}\\
\cline{2-15}
~& \multicolumn{1}{|c}{Simulator} & \multicolumn{3}{|c}{Isaac Sim} & \multicolumn{4}{|c}{AI2-THOR} & \multicolumn{3}{|c}{Habitat} & \multicolumn{1}{|c}{SAPIEN} & \multicolumn{2}{|c|}{Matterport}  \\
\hline
% \multicolumn{1}{|c|}{\multirow{4}*{\rotatebox{90}{Data}}} & \multicolumn{1}{|c|}{Room} &  \\
% \multicolumn{1}{|c|}{~}& \multicolumn{1}{|c|}{Building} & 100\\
\multicolumn{1}{|c|}{\multirow{3}*{\rotatebox{90}{Scene}}}& \multicolumn{1}{|c|}{Scene Types} & \textbf{89} & 8 & 1 & 1 & 1 & 1 & 1 & 1 & 1 & 1 & 1 & 1 & 1\\
\multicolumn{1}{|c|}{~}& \multicolumn{1}{|c|}{City Scale } &  \CheckMark & \CrossMark & \CrossMark  & \CrossMark  & \CrossMark  & \CrossMark  & \CrossMark  & \CrossMark  & \CrossMark  & \CrossMark  & \CrossMark  & \CrossMark  & \CrossMark  \\
\multicolumn{1}{|c|}{~}& \multicolumn{1}{|c|}{Region Label} &  \CheckMark & \CrossMark &  \CheckMark & \CheckMark & \CheckMark & \CheckMark & \CheckMark & \CrossMark & \CheckMark & \CheckMark & \CrossMark & \CrossMark & \CrossMark \\
\hline 
\multicolumn{1}{|c|}{\multirow{4}*{\rotatebox{90}{Object}}} & \multicolumn{1}{|c|}{Interactive} &  \CheckMark & \CheckMark & \CheckMark & \CheckMark & \CheckMark & \CheckMark & \CheckMark & \CrossMark & \CheckMark & \CheckMark & \CheckMark & \CrossMark & \CrossMark \\
% \multicolumn{1}{|c|}{~} & \multicolumn{1}{|c|}{Non-interactive object} &  22k \\
\multicolumn{1}{|c|}{~}& \multicolumn{1}{|c|}{Part Label} & \CheckMark  & \CrossMark & \CrossMark & \CrossMark & \CrossMark & \CrossMark & \CrossMark & \CrossMark & \CrossMark & \CrossMark & \CheckMark & \CrossMark & \CrossMark\\
\multicolumn{1}{|c|}{~}& \multicolumn{1}{|c|}{Material Label} & \CheckMark & \CheckMark & \CrossMark & \CrossMark  & \CrossMark  & \CrossMark  & \CrossMark  & \CrossMark  & \CrossMark  & \CrossMark  & \CrossMark  & \CrossMark  & \CrossMark \\
\multicolumn{1}{|c|}{~}& \multicolumn{1}{|c|}{Language Caption} & \CheckMark  & \CrossMark & \CrossMark  & \CrossMark  & \CrossMark  & \CrossMark  & \CrossMark  & \CrossMark  & \CrossMark  & \CrossMark  & \CrossMark  & \CrossMark  & \CrossMark\\
\hline
\multicolumn{1}{|c|}{\multirow{4}*{\rotatebox{90}{Platform}}} & \multicolumn{1}{|c|}{Learning-based Controller} & \CheckMark & \CrossMark & \CrossMark  & \CrossMark & \CrossMark & \CrossMark & \CrossMark & \CrossMark & \CrossMark & \CrossMark & \CheckMark & \CrossMark & \CrossMark\\
% \multicolumn{1}{|c|}{~} & \multicolumn{1}{|c|}{RL-based controller} &  & - \\
\multicolumn{1}{|c|}{~} & \multicolumn{1}{|c|}{LLM NPC} & \CheckMark & \CrossMark & \CrossMark & \CrossMark & \CrossMark & \CrossMark & \CrossMark & \CrossMark & \CrossMark & \CrossMark & \CrossMark & \CrossMark & \CrossMark  \\
\multicolumn{1}{|c|}{~} & \multicolumn{1}{|c|}{Kinematics} & \CheckMark & \CheckMark & \CheckMark & \CrossMark & \CrossMark & \CrossMark & \CrossMark & \CrossMark & \CrossMark & \CrossMark & \CheckMark & \CrossMark & \CrossMark \\
\multicolumn{1}{|c|}{~} & \multicolumn{1}{|c|}{Continuous} & \CheckMark & \CheckMark & \CheckMark & \CrossMark & \CrossMark & \CrossMark & \CrossMark & \CheckMark & \CheckMark & \CheckMark & \CheckMark & \CrossMark & \CrossMark\\
\hline
\multicolumn{1}{|c|}{\multirow{5}*{\rotatebox{90}{Benchmark}}}  & \multicolumn{1}{|c|}{Language Instruction} & \CheckMark & \CrossMark & \CheckMark & \CheckMark & \CrossMark & \CrossMark & \CrossMark & \CheckMark & \CrossMark & \CheckMark & \CheckMark & \CheckMark & \CheckMark \\
\multicolumn{1}{|c|}{~} & \multicolumn{1}{|c|}{Task Generation} & \CheckMark & \CrossMark & \CrossMark & \CrossMark & \CrossMark & \CrossMark & \CrossMark & \CrossMark & \CrossMark & \CrossMark & \CrossMark & \CrossMark & \CrossMark \\
\multicolumn{1}{|c|}{~} & \multicolumn{1}{|c|}{Navigation} & \CheckMark & \CheckMark & \CrossMark & \CheckMark & \CrossMark & \CrossMark & \CrossMark & \CheckMark & \CheckMark & \CheckMark & \CrossMark & \CheckMark & \CheckMark \\
\multicolumn{1}{|c|}{~} & \multicolumn{1}{|c|}{Social Interaction} & \CheckMark & \CrossMark & \CrossMark & \CrossMark & \CrossMark & \CrossMark & \CrossMark & \CrossMark & \CrossMark & \CheckMark & \CrossMark & \CrossMark & \CheckMark \\
\multicolumn{1}{|c|}{~} & \multicolumn{1}{|c|}{Manipulation} & \CheckMark & \CheckMark & \CheckMark & \CheckMark & \CheckMark & \CheckMark & \CrossMark & \CrossMark & \CheckMark & \CheckMark & \CheckMark & \CrossMark & \CrossMark \\
\hline
\end{tabular}
} % \resizebox
\vspace{-15pt}
\label{tab:platform_comparison}
\end{table}
\noindent\textbf{Embodied AI Benchmarks.} 
VirtualHome~\cite{puig2018virtualhome} and Alfred~\cite{alfred} abstract physical interactions to focus on symbolic reasoning but lack physical realism and action scope. Habitat~\cite{savva2019habitat} employs 3D scans of real houses for navigation tasks~\cite{batra2020objectnav} but lacks physics-based interaction. To enhance physical realism, Habitat 2.0~\cite{homerobot} and iGibson~\cite{shen2021igibson, li2021igibson2} offer realistic actions, environment interactions, and object state simulations. Recent simulation platforms such as ManiSkills~\cite{gu2023maniskill2}, TDW~\cite{tdwworld}, SoftGym~\cite{softgym}, and RFUniverse~\cite{fu2022rfuniverse} focus on physical realism yet lack task diversity. To enhance task diversity, some works explore language-conditioned tasks~\cite{raven, james2019rlbench, mees2022calvin}. GenSim~\cite{wang2023gensim} explores the potential of LLMs in generating table-top tasks using code scripts. RoboGen~\cite{wang2023robogen} and MimicGen~\cite{mandlekar2023mimicgen} leverage generative models and LLMs to generate tasks. In comparison, our platform supports automatically AI-generated tasks ranging from navigation to mobile manipulation across diverse and higher-quality room layouts. Moreover, our benchmark designs include social interactions~\cite{socialnavigation, puig2018virtualhome} between agents and human-like NPCs with access to simulator knowledge. In terms of scene diversity, previous benchmarks are mainly established in the home environments. Compared with Behavior-1K~\cite{behavior} offering 8 different scene types, GRUtopia provides a more diverse dataset encompassing 89 scene categories and builds a city-scale 3D scene with those assets. A detailed comparison between GRUtopia and other benchmarks is shown in Tab.~\ref{tab:platform_comparison}. 

\noindent\textbf{Low-level control policy.} Low-level control for legged robots~\cite{agileleggedrobot, walkinminutes} is typically trained in simulators~\cite{todorov2012mujoco, raisimLib, webots}. Isaac Orbit~\cite{mittal2023orbit} and Gym~\cite{walkinminutes} enable massive parallel training but are limited to specific terrains. In terms of low-level control policy for manipulation, the Aloha series~\cite{aloha, mobilealoha} demonstrates impressive imitation learning in bimanual and mobile manipulation. Recent works like RT-1~\cite{rt12022arxiv} scale robotic learning across tasks, while BC-Z~\cite{bcz} generalizes to unseen skills from human videos. Our simulator provides more diverse environments with physical simulations, on which policy training can be performed. Besides,  it also offers low-level locomotion and manipulation policy APIs, to foster the study of high-level task planning.

\noindent\textbf{NPC in Simulator.} Recent advancements emphasize social interaction as pivotal in human-robot interaction. Habitat 3.0~\cite{socialnavigation} explores collaborations between humanoid and robot agents in home settings, similar to generative agents~\cite{generativeagent} that use LLMs to simulate authentic human behaviors. 
Not limited to task assignment, the NPC design in our research can provide crucial information during task execution, making it transcend traditional human-in-the-loop paradigm. 
Specifically, our platform employs LLM-driven NPCs to assist and interact with agents to complete domestic tasks, harnessing environmental data and utilizing platform APIs similar to MineDoJo~\cite{fan2022minedojo}.

\begin{figure*}
% \vspace{-1ex}
     \centering
          \begin{subfigure}[b]{0.32\textwidth}
         \centering
         \includegraphics[width=\textwidth]{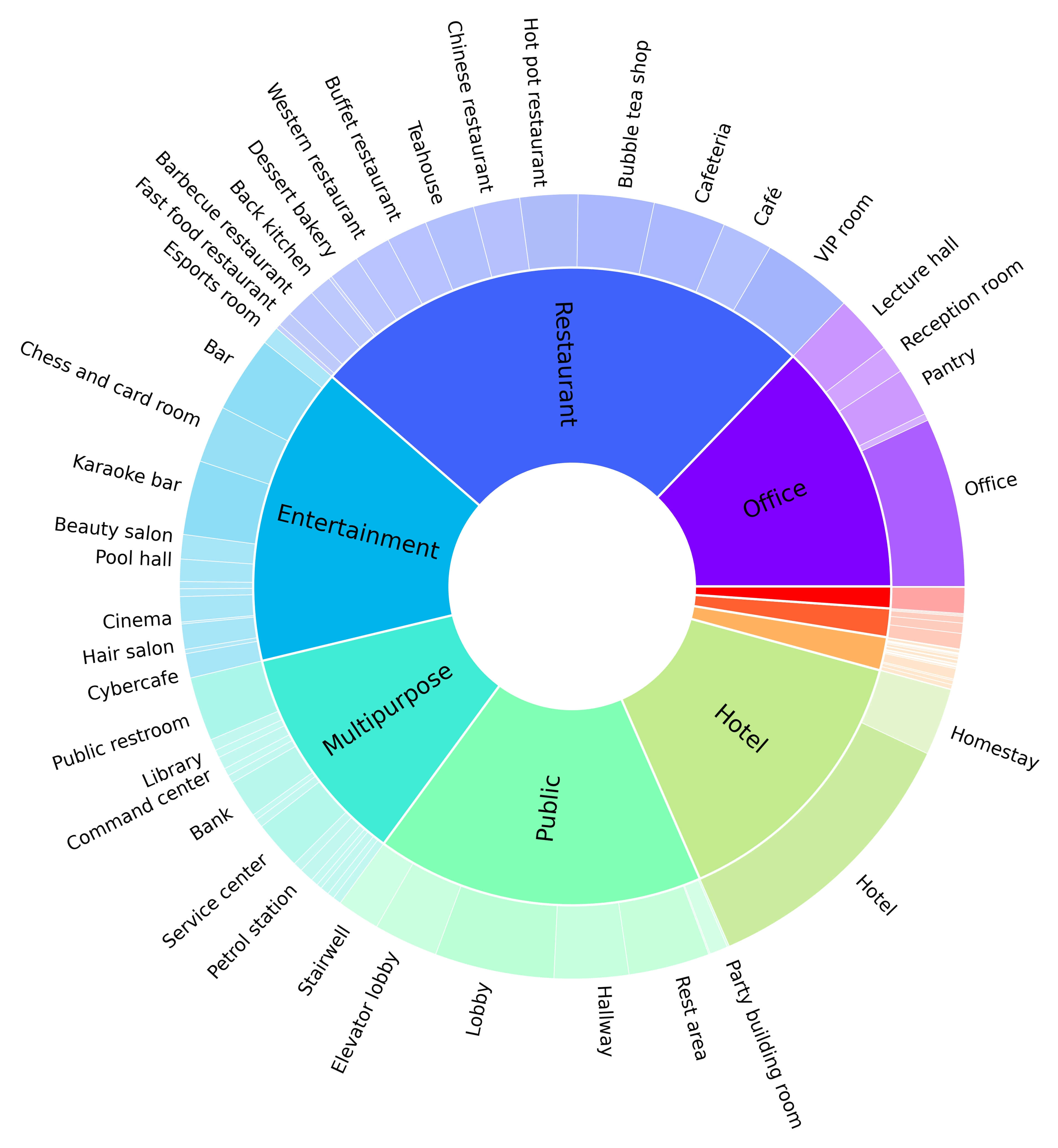}
         \caption{Scene Category Statistics.}
         \label{fig:scene-stat} 
     \end{subfigure}
     \hfill
     % \begin{subfigure}[b]{0.64\textwidth}
     %     \centering
     %     \includegraphics[width=\textwidth]{figures/interactive_objs.jpg}
     %     \caption{Interactive Objects Statistics with Part Annotations.}
     %     \label{fig:instance-stat} 
     % \end{subfigure}
     % \\[0.05in]
     % \begin{subfigure}[b]{0.34\textwidth}
     %    \centering
     %    \includegraphics[width=\textwidth]{figures/curated_scenes.jpg}
     %    \caption{Regions Statistics.}
     %    \label{fig:3d-box-prompt-stat} 
     % \end{subfigure}
     \begin{subfigure}[b]{0.66\textwidth}
         \centering
         \includegraphics[width=\textwidth]{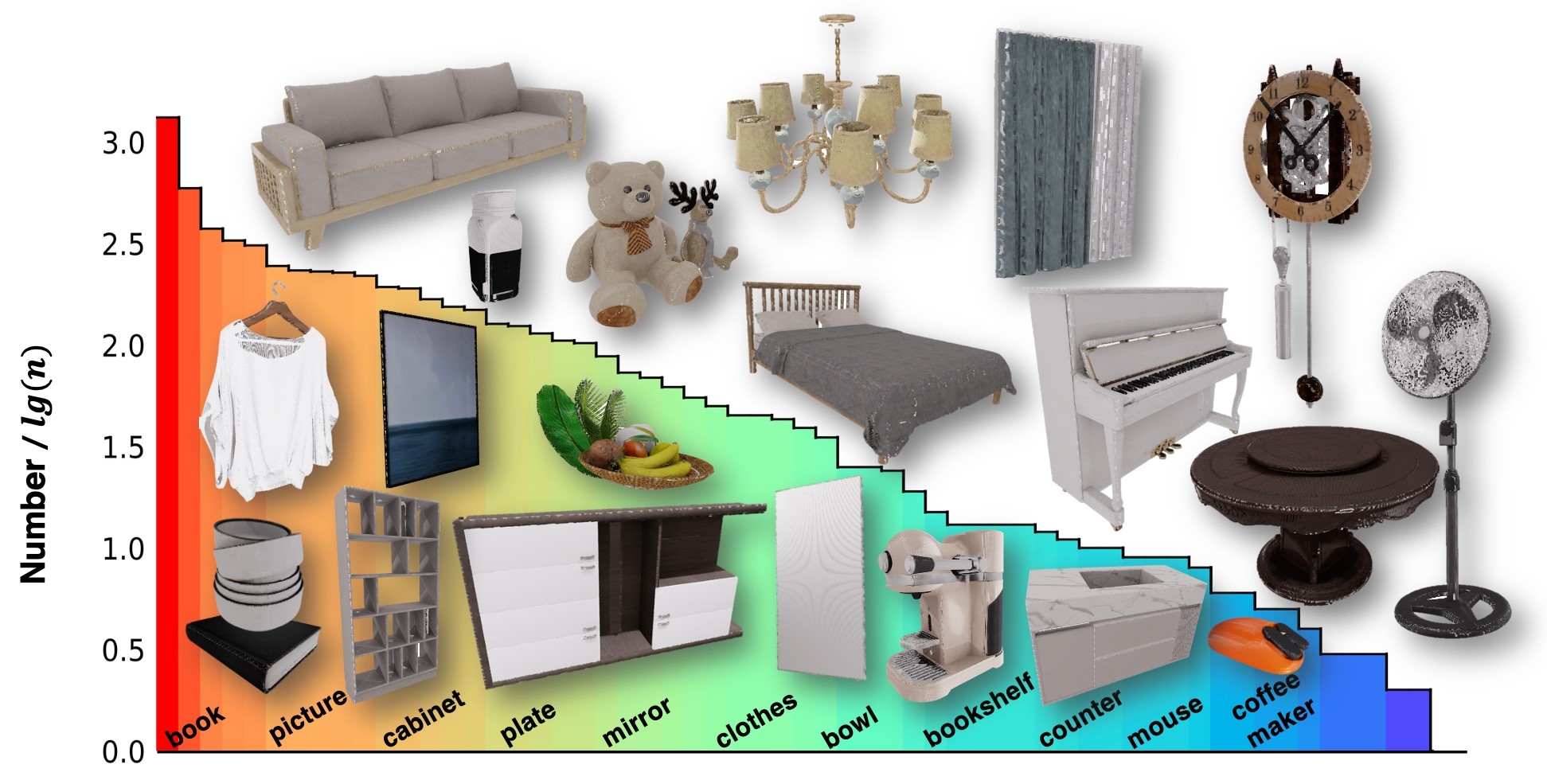}
         \caption{Objects Statistics.}
         \label{fig:obj-stat} 
     \end{subfigure}
     \caption{The richness and diversity of scenes and objects in GRScenes. (a) The distribution of scenes is shown in a fine granularity. GRScenes covers a vast range of functional scene categories. (b) The spectrum of annotated objects and some examples present in GRScenes.} 
     \label{fig:dataset-stat}
     % \vspace{-5ex}
\end{figure*}

% TODO

% \section{Bridging the Gap between Embodied AI and Physical World}
\vspace{-2ex}
\section{Dataset \& Platform}
\vspace{-1.5ex}
\label{sec:utopia}
In this section, we first introduce the GRUtopia platform, which is built on a large-scale 3D scene dataset—GRScenes—featuring diverse objects (Sec.\ref{sec:grscenes}). We then describe GRResidents, the NPC system driven by LLMs for social interactions (Sec.\ref{sec: generative npc in the wild}), and the World Knowledge Manager (WKM), which provides platform APIs for NPCs. In Sec.~\ref{sec: control api}, we provide the control APIs for various types of robots and evaluate the stability of locomotion in GRScenes.

% Overall a table to compare GRUtopia vs. previous platforms
\vspace{-1ex}
\subsection{GRScenes: Fully Interactive Environment at Scale}\label{sec:grscenes}
\vspace{-1ex}

To build a platform for training and benchmark embodied agents, fully interactive environments with diverse scene and object assets are indispensable. Therefore, we collect a large-scale 3D synthetic scene dataset with diverse object assets, which serves as the foundation of our GRUtopia platform.
Next, we will elaborate on its details from three aspects: scenes, objects, and the corresponding hierarchical multi-modal annotations.

\noindent\textbf{Diverse and Realistic Scenes.}
Due to the limited amount and categories of open-source 3D scene data, we first collect about 100k high-quality synthetic scenes from designer websites to obtain diverse scene prototypes. These prototypes are then cleaned, annotated with semantics in the region and object level, and finally combined together to form towns as the fundamental playground for robots.
As shown in Fig.~\ref{fig:dataset-stat}-(a), except for the common home scenes, our dataset consists of $\sim$30\% scenes from other diverse categories, such as restaurants, offices, public, hotels, entertainment, \emph{etc.}
From the large-scale dataset, we preliminarily curate 100 scenes with fine-grained annotations for the open-source benchmark.
These 100 scenes contain 70 home and 30 commercial scenes, among which the home scenes are composed of comprehensive common regions with diverse areas (see statistics in the appendix), and the commercial scenes cover common types, including hospitals, supermarkets, restaurants, schools, libraries, and offices.
Furthermore, we collaborate with several professional designers to distribute objects following human habits to make these scenes more realistic, as shown in Fig.~\ref{fig:example}, which are usually ignored in previous works~\cite{puig2018virtualhome,shen2021igibson,ai2thor}.
In this way, we establish a solid foundation to enable general robots to train and test in realistic, large-scale environments.

\noindent\textbf{Interactive Objects with Part-Level Annotations.}
These scenes contain several 3D objects originally, but some of them do not have internal modeling, making it infeasible to train robots to interact with these objects. To address this problem, we work with a professional team to modify these assets and create complete objects to make them interactable in a physically plausible manner.
In addition, to provide more comprehensive information to enable agents to interact with these assets, we annotate all the objects with fine-grained part labels attached to the interactive parts in X-form in NVIDIA Omniverse.
Finally, the 100 scenes contain 2,956 interactive objects and 22,001 non-interactive objects from 96 categories, whose distributions are shown in Fig.~\ref{fig:dataset-stat}-(b)\footnote{For the remaining scenes, we plan to develop a method to automatically construct the interior structure of objects in the future}.

\noindent\textbf{Hierarchical Multi-modal Annotations.} Finally, to enable multi-modal interaction of embodied agents and the environment as well as NPCs, we also need language annotations for these scenes and objects.
In contrast to previous multi-modal 3D scene datasets~\cite{scanrefer,referit3d,scanqa} only focusing on the object level or inter-object relationships, we also take different granularities of scene elements into consideration, such as object-region relationships.
Given the lack of region labels, we first design a UI to annotate regions on a bird's eye view of the scene with polygons and then can involve object-region relationships in language annotations.
For each object, we prompt powerful VLMs (\emph{e.g.}, GPT-4v) with rendered multi-view images to initialize the annotation followed by human manual checks.
The resulting language annotations provide the foundation to generate embodied tasks for subsequent benchmarks in Sec.~\ref{sec: gr-bench}.
See more details in the supplementary material.
\subsection{GRResidents: Generative NPCs in 3D Environments}
\label{sec: generative npc in the wild}
% \vspace{-1ex}

\begin{figure}
    \centering
    \includegraphics[width=\textwidth]{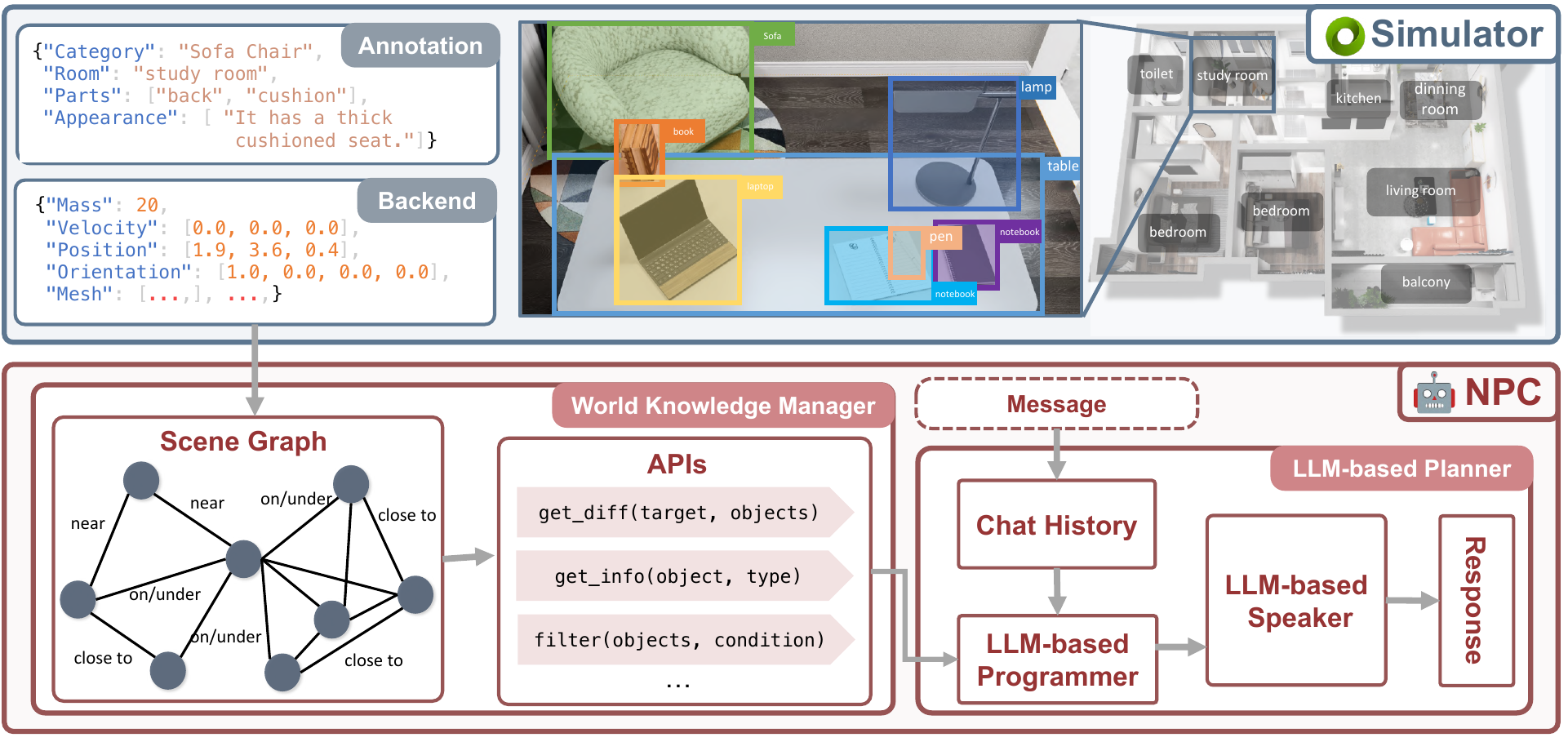}
    % \vspace{-2ex}
    \vspace{-0.05in}
    \caption{Overview of GRResidents (Sec.~\ref{sec: generative npc in the wild}). It comprises two modules: (a) A world knowledge manager that organizes scene knowledge from dataset annotations and simulator backend and provides APIs for knowledge retrieving. (b) An LLM planner that is able to retrieve global knowledge from the world knowledge manager and generate responses according to dialogue context.}
    % \vspace{-5ex}
    \label{fig:npc_overview}
\end{figure}

In GRUtopia, we endowed the world with social capabilities by embedding some ``residents'', \ie generative NPCs driven by LLMs, thereby simulating social interactions within an urban environment. This NPC system is named as GRResidents. One of the main challenges of building authentic virtual characters in 3D scenarios is integrating the ability of 3D perception. However, the virtual characters can easily access scene annotations and internal states of the simulated world, making robust perception achievable. In this way, we devise a World Knowledge Manager (WKM) that manages the dynamic knowledge of the real-time world state and provides access through a series of data interfaces. Resorting to WKM, the NPCs can retrieve desired knowledge and perform fine-grained object grounding through parameterized function calls, which form the core of their perception capabilities. 

\noindent\textbf{World Knowledge Manager (WKM).} The main duty of WKM is managing knowledge of the virtual environment persistently and offering high-level knowledge of the scene to NPCs. Concretely, the WKM takes hierarchical annotations and scene knowledge from our dataset and simulator backend respectively to build the scene graph as the scene representation, where each node represents an object instance and the edges indicate the spatial relationship between the objects. We adopt the spatial relationships defined in Sr3D~\cite{referit3d} as the relation space. This scene graph is preserved by WKM at each simulation step. Moreover, WKM exposes three core data interfaces to extract knowledge from the scene graph: 1) \texttt{\funcc{find\_diff(}\argc{target}, \argc{objects}\funcc{)}}: compare the difference between an \texttt{\argc{target}} object and a set of other \texttt{\argc{objects}}, 2) \texttt{\funcc{get\_info(}\argc{object}, \argc{type}\funcc{)}}: get knowledge of an \texttt{\argc{object}} in terms of the desired attribute \texttt{\argc{type}}, and 3) \texttt{\funcc{filter(}\argc{objects}, \argc{condition}\funcc{)}}: filtering the \texttt{\argc{objects}} according to the \texttt{\argc{condition}}. Please refer to supplementary materials for more details.

\noindent\textbf{LLM Planner.} The decision-making module of NPCs is an LLM-based planner that comprises three components (Fig.~\ref{fig:npc_overview}): a memory module that stores the chat history between the NPC and other agents, an LLM programmer that uses interfaces from WKM to query scene knowledge, and an LLM speaker that consumes the chat history and queried knowledge to produce responses. When an NPC receives a message, it first stores the message in memory and forwards the updated history to the LLM programmer. The programmer then iteratively calls the data interfaces to query the necessary scene knowledge. Finally, the knowledge and history are sent to the LLM speaker, which generates the response.

\begin{wrapfigure}[8]{r}{7cm}
\vspace{-5pt}
\begin{minipage}{0.5\textwidth}

% \centering
\renewcommand\arraystretch{1.1}
\captionof{table}{\small Cross-verification (Referring \& Grounding) accuracy (\%) and QA score of our NPCs driven by different LLMs.\label{tab:npc}} % \caption
\resizebox{\linewidth}{!}{ %< auto-adjusts font size to fill line
\begin{tabular}{@{}|cccc|@{}}
\hline
\multicolumn{1}{|c|}{LLM} & \multicolumn{1}{c|}{Referring} & \multicolumn{1}{c|}{Grounding} & \multicolumn{1}{c|}{QA}\\
\hline
\multicolumn{1}{|c|}{GPT-4o\cite{gpt4o}} & 100.0 & 93.2 & 95.8\\
\multicolumn{1}{|c|}{InternLM-2-Chat\cite{cai2024internlm2}} & 95.9 & 83.3 & 88.7\\
\multicolumn{1}{|c|}{Llama-3-70B\cite{llama3}} & 100.0 & 88.6 & 92.5\\
\hline
\end{tabular}
} % \resizebox

\end{minipage}
\end{wrapfigure}

\noindent\textbf{Experiments.} We conduct experiments on object referring, language grounding, and object-centric QA to demonstrate that our NPCs can: (1) generate object captions, (2) locate objects by description, and (3) provide object information for agents. The backend LLMs of NPC in these experiments include GPT-4o~\cite{gpt4o}, InternLM2-Chat-20B~\cite{cai2024internlm2}, and Llama-3-70B-Instruct~\cite{llama3}.

\begin{wrapfigure}[19]{r}{7cm}
% \vspace{-5pt}
\begin{minipage}{0.5\textwidth}

\includegraphics[width=\textwidth]{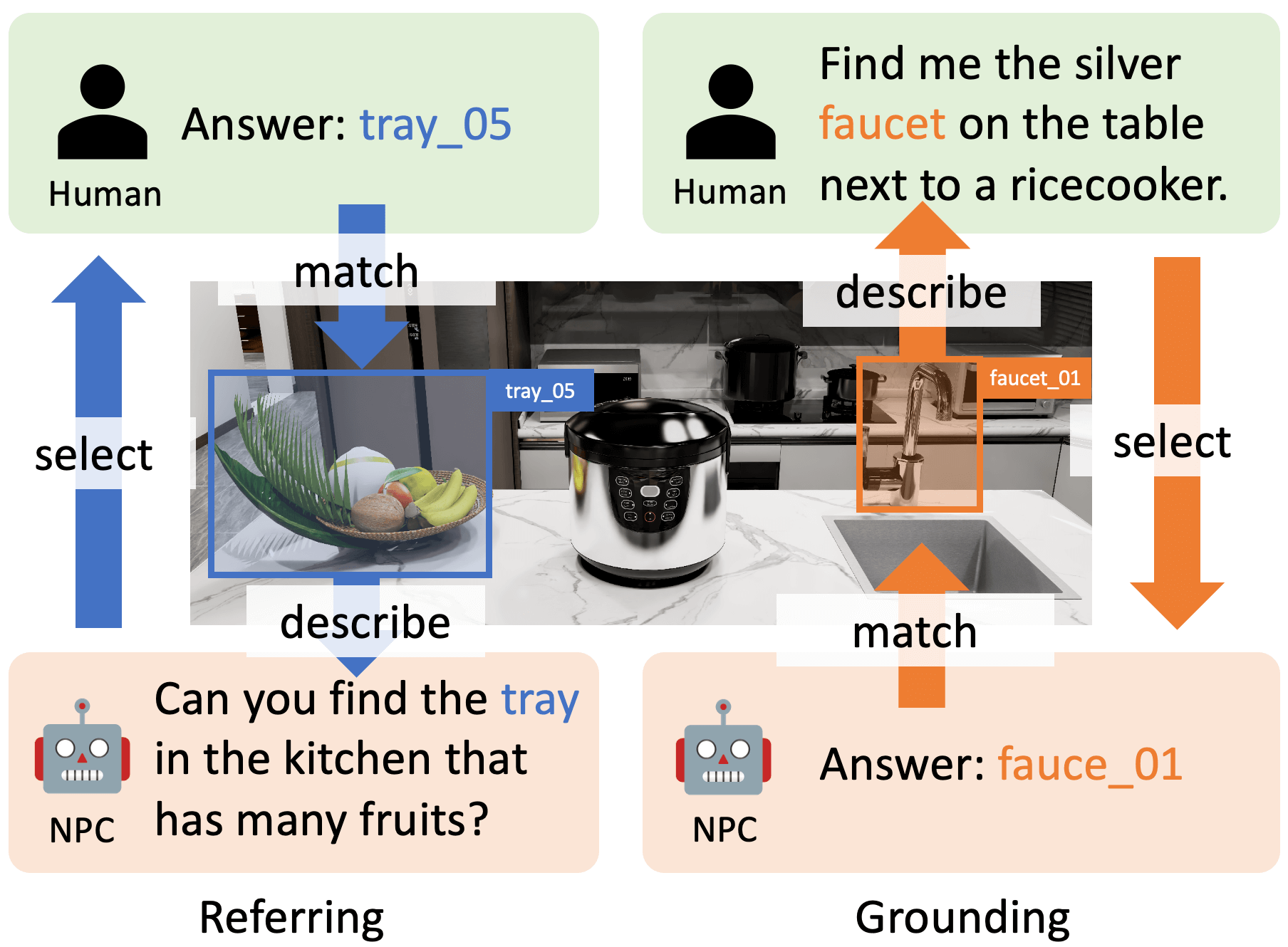}
\captionof{figure}{\small Evaluation of the NPC's capabilities in description and grounding through cross-verification between NPC and human.\label{fig:cross_veri}
} % \caption

\end{minipage}
\end{wrapfigure}

As shown in Fig.~\ref{fig:cross_veri}, in the referring experiment, we employ a human-in-the-loop evaluation. The NPC randomly selects an object and describes it, then the human annotator selects an object according to the description. Referring is successful if the human annotator can locate the correct object corresponding to the description. In the grounding experiment, the role of the human annotator is played by GPT-4o\cite{gpt4o}, which provides an object description, and the NPC locates it. Grounding is successful if the NPC can locate the corresponding object.
The success rate in Tab.~\ref{tab:npc} (Referring and Grounding) shows that different LLMs achieve 95.9\%-100\% and 83.3\%-93.2\% accuracy, validating the accuracy of our NPC framework in object referring and grounding across LLMs.

In the object-centric QA experiment, we evaluate the NPC's ability to provide object-level information to agents through question-answering in navigation tasks. 
We design a pipeline that generates object-centric navigation episodes simulating real-world scenarios where the agent asks the NPC questions for information and takes actions based on the answers. 
Given the agents' questions, we assess the NPC based on the semantic similarity between its answers and the ground-truth answers. The overall score, as shown in Table~\ref{tab:npc} (QA), demonstrates that our NPC can provide precise and helpful navigation assistance. Please refer to supplementary materials for more details.

\subsection{Robot Control APIs}\label{sec: control api}
As physical simulation requires collision handling, low-level control APIs are essential for managing robot agents within the simulator. Unlike prior works that use animation and set positions for pseudo-action execution, we present RL-based controllers as APIs for driving robots. This methodology facilitates the deployment of agent algorithms in executing high-level tasks by leveraging pre-trained, robust low-level control systems. Specifically, we have developed and provided locomotion policies as APIs following state-of-the-art policy learning practices~\cite{long2023him,liu2024visual} tailored to various robots, including humanoid robots (Unitree H1, Unitree G1, Fourier GR-1) and quadrupeds (Unitree Aliengo with Unitree Z1 Arm). These locomotion APIs have been meticulously designed to be user-friendly, enabling researchers and developers to integrate sophisticated control mechanisms without delving into the intricacies of RL training.

We conduct extensive evaluations of the locomotion controllers on both Unitree H1 and Unitree Aliengo (with Z1 arm) in two distinct environments: an open flat terrain and a cluttered, furniture-rich environment. We task the robots with point-to-point navigation using a simple shortest path search for path planning. A significant performance drop in success rates is observed, from 100\% to 58\% for H1 and 14\% for Aliengo (See Tab.~\ref{tab:APIs}).  Similarly, we test the manipulation capability and noted a reduction in success rate from 100\% to 8\%. We found that this disparity is mainly caused by the failure of path or motion planning, despite the robust performance of low-level control. This highlights the challenges of deploying policies trained in simple environments to real-world complex scenarios. Consequently, we advocate for a more integrated research approach that combines low-level control studies with high-level task execution in these realistic settings. As a temporary workaround, we introduce a \textit{reset} function that restores the agent's stable kinematic state and uses the number of resets to evaluate the robustness of the low-level policies. In Sec.~\ref{sec:experiments}, we study the applications and limitations of these control APIs in handling high-level tasks.

\begin{table}[t]
\centering
\renewcommand\arraystretch{1.1}
\caption{\small Performance comparison of different controllers. (TE: Trajectory Error (m), SR: Success Rate (\%), AS: Average Speed (m/s), AT: Average Time (s), PE: Position Error of the end effector (m).)
} % \caption
\resizebox{\linewidth}{!}{ %< auto-adjusts font size to fill line
\begin{tabular}{@{}c|cccc|cccc|cc|cc|@{}}
\hline
\multicolumn{1}{|c|}{\multirow{2}*{Embodiments}}& \multicolumn{4}{c|}{Locomotion (Flat)} & \multicolumn{4}{c|}{Locomotion (House)} & \multicolumn{2}{c|}{Manip. (Flat)} & \multicolumn{2}{c|}{Manip. (House)}\\
\multicolumn{1}{|c|}{~} & TE & SR & AS & AT & TE & SR & AS & AT & PE & SR  & PE & SR \\
\hline
\multicolumn{1}{|c|}{Unitree H1} & 0.01 & 100 & 0.19 & 43.51 & 0.07 & 58 & 0.17 & 36.46 & - & - & - & - \\
\multicolumn{1}{|c|}{Unitree Aliengo+Z1} & 0.01 & 100 & 0.25 & 32.88 & 0.13 & 14 & 0.16 & 32.35 & 0.00 & 100 & 0.56 & 8 \\
\hline
\end{tabular}
} % \resizebox

\label{tab:APIs}
\end{table}

% TODO

\vspace{-2ex}
\section{GRBench: A Benchmark for Assessing Embodied Agents}\label{sec: gr-bench}
\vspace{-1.5ex}
An embodied agent is expected to actively perceive its environment, engage in dialogue to clarify ambiguous human instructions, and interact with its surroundings to complete tasks. In this section, we introduce our benchmarks in GRUtopia, including the three benchmark setups in Sec.\ref{sec: benchmark setup} and the respective evaluation metrics for navigation and manipulation tasks in Sec.\ref{sec: task evaluation}.

\begin{figure}[h]
\begin{center}
\vspace{-1ex}
\includegraphics[width=\linewidth]{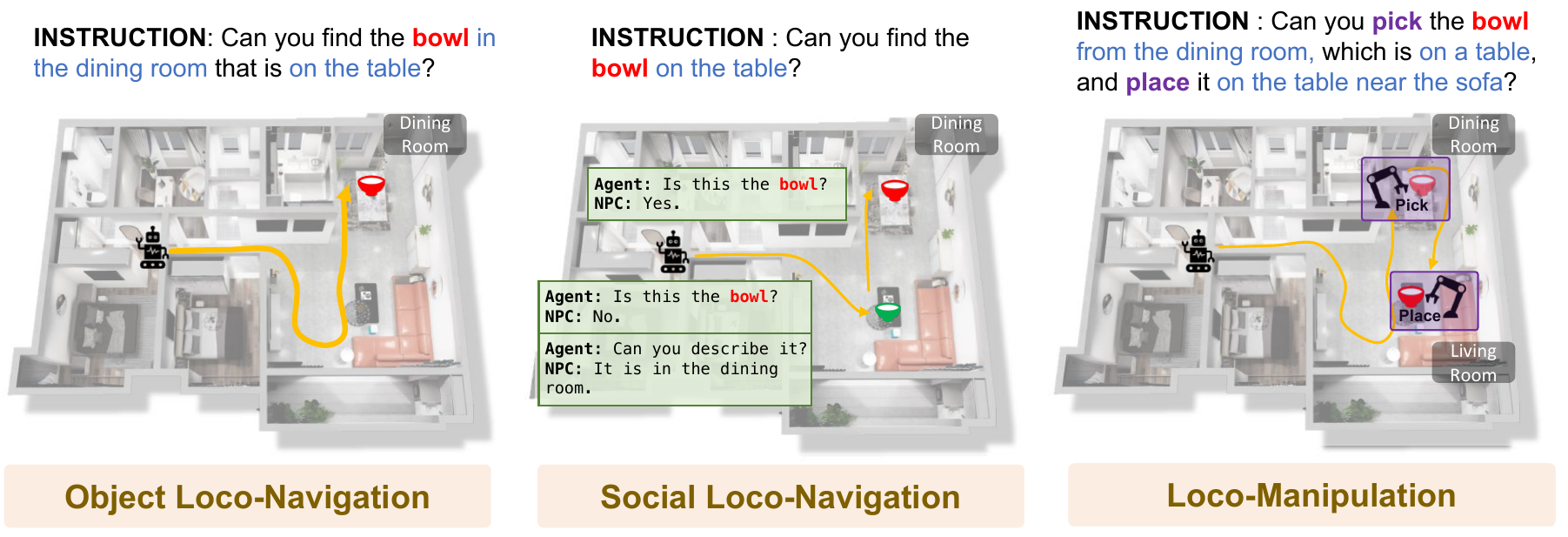}
\end{center}
\vspace{-0.08in}
\caption{GRBench task examples. Three benchmarks are established for evaluating embodied agents: \textit{Object Loco-Navigation}, \textit{Social Loco-Navigation}, and \textit{Loco-Manipulation}. The {\color{red} target object} in the instruction are subject to some \textcolor[RGB]{68,114,169}{constraints} generated by the world knowledge manager. \textcolor[RGB]{255,192,0}{Navigation paths}, \textcolor[RGB]{169,209,142}{dialogues}, and \textcolor[RGB]{112,48,160}{actions} are depicted in the figure; please zoom in for better visualization.
}
% \vspace{-3ex}
\label{fig:benchmark example}
\end{figure}

\vspace{-1ex}
\subsection{Benchmark Setups}\label{sec: benchmark setup}
\vspace{-1ex}
As shown in Fig.~\ref{fig:benchmark example}, which displays some example cases, we set up three benchmarks for the comprehensive evaluation of an embodied agent: 1) \textit{Object Loco-Navigation}, which assesses active perception and navigation; 2) \textit{Social Loco-Navigation}, which evaluates effective communication with NPCs for instruction clarification; and 3) \textit{Loco-Manipulation}, which measures mobile manipulation. We generate 300 episodes for each benchmark (100 for validation and 200 for the testing set for embodied agents). Refer to the supplementary files for more implementation details of task definition and task generation.

\noindent\textbf{Benchmark 1: Object Loco-Navigation.} The object loco-navigation task requires the agent to navigate to the target object based on a given language goal. The world knowledge manager (Sec.~\ref{sec: generative npc in the wild}) ensures the target is uniquely identified using non-ambiguous natural language. An episode is successful if the target object appears in the agent’s field of view.

\noindent\textbf{Benchmark 2: Social Loco-Navigation.} Recognizing that human intention may not always be clearly provided, the social loco-navigation task evaluates the agent’s ability to actively interact with NPCs and identify the target through dialogue. The agent must ask questions to clarify the features of the target objects (with a maximum of three dialogue rounds). As with Object Loco-Navigation, an episode is considered successful if the target object appears in the agent’s field of view.

\noindent\textbf{Benchmark 3: Loco-Manipulation.} The loco-manipulation task builds on loco-navigation by testing a robot’s ability to pick up and place objects using its arm. This task involves arm manipulations to pick up the target handheld object from initial positions and place it in the target receptacle in the correct position. Agents are required to understand both the appearance and relationships between objects and the receptacles. This task is defined by at most two conditions that describe the target location of the handheld object, including its appearance and relationship with the target receptacles; thus, multiple solutions are allowed since both the handheld objects and target receptacles are not guaranteed to be unique. An episode is considered successful if the handheld object is finally placed in a position that satisfies all conditions in the task specifications. This makes this task more challenging as it requires accurately transporting the correct object to the target position. 

\noindent\textbf{Robot Setups.} All episodes begin with the agent at a predetermined location and orientation in the task specification. For navigation (Benchmark 1 \& 2), we use a Unitree H1 humanoid robot equipped with an RGB-D camera for perception. For mobile manipulation (Benchmark 3), since the Unitree H1 currently lacks sufficient manipulation capabilities, the setup uses a mobile manipulator combining an AlienGo as the mobile base and a Unitree Z1 as the manipulator. An RGB-D camera is mounted on a pole located $0.8$ meters above the AlienGo to ensure the camera’s FoV for environmental perception.

\subsection{Baselines}
\noindent\textbf{Zero-Shot VLM Baselines.} We adopt state-of-the-art VLMs from open-source representatives such as InternVL-chat-1.5 \cite{chen2024far}, as well as closed-source ones from GPT-4o\cite{gpt4o} and Qwen-VL\cite{Qwen-VL} families. For \textbf{Object Loco-Navigation}, we directly input the current image observation of the robot and the language prompt into the VLM. The VLM then chooses one action from the following 12 actions as output: (1) \textit{Move forward 2/4/6 meters}; (2) \textit{Advance 2/4/6 meters diagonally to the left/right}; (3) \textit{Turn left/right 90 degrees}; (4) \textit{Stop}. After the robot executes the chosen action, new observation and language prompt will be given to the VLM to choose the next action. This process will continue until the action output by the VLM is \textit{Stop}. For \textbf{Social Loco-Navigation}, a new action, \textit{Ask}, is introduced. This allows the robot to ask NPC for more information about the target object. For \textbf{Loco-Manipulation}, new action types \textit{Pick} and \textit{Place} are introduced to support the mobile manipulation ability of the robot agent. Please refer to the appendix for more details.

\begin{figure}[t]
    \centering
    \includegraphics[width=1\linewidth]{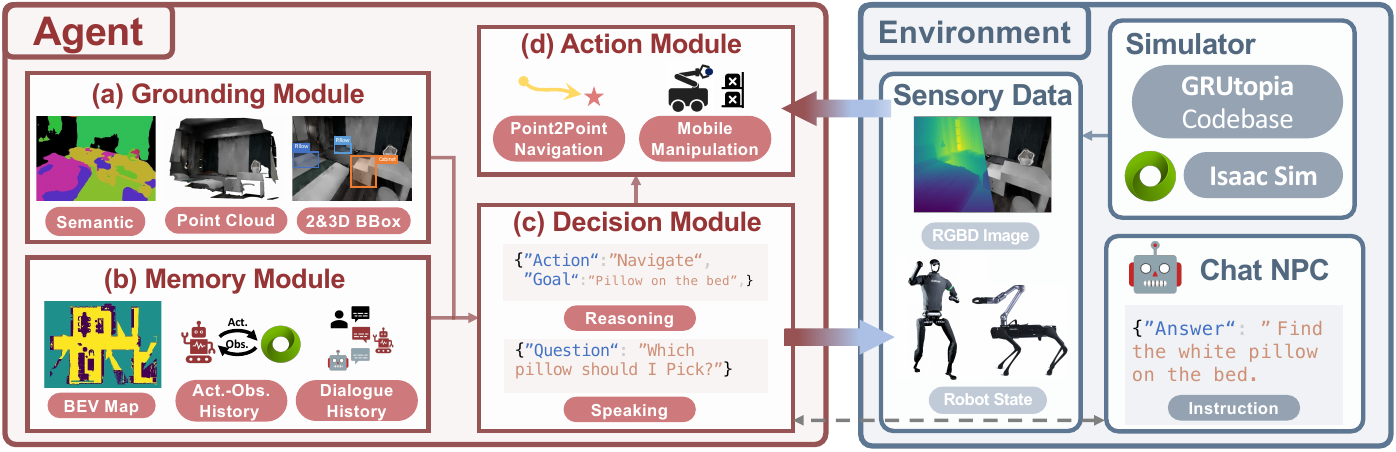}
    \caption{Overview of the baseline agent. The grounding module (a) processes raw sensory data into semantic-rich information, and the memory module (b) stores historical information like action-observation history. The decision module (c), which consists of a VLM or LLM, makes action decisions based on information from (a) and (b), while the action module (d) executes the output action. The environment simulates the physical changes resulting from the action and produces sensory data. The agent can also choose to ask the oracle NPC for further instructions about the task.}
    \label{fig:agent_overview}
\end{figure}

\noindent\textbf{LLM Agent Baselines.} As shown in Fig.~\ref{fig:agent_overview}, the proposed LLM agent consists of a grounding module, a memory module, a decision module, and an action module. The environmental inputs to the agent are egocentric observations of the agent and the current state of the robot. Through the collaborative interactions among these modules, the agent can effectively analyze and utilize environmental inputs, enabling it to engage in both physical and linguistic interactions with the environment. Please refer to the appendix for more details.

\vspace{-1.5ex}
\subsection{Evaluation}\label{sec: task evaluation}
\vspace{-1.5ex}
In this subsection, we outline the criteria and metrics used to evaluate the performance of the agent across different benchmarks.

\noindent\textbf{Success Criteria.}
For Benchmarks 1 and 2, an episode is considered successful if the target object is within the agent’s field of view and less than 3 meters away. For Benchmark 3, success is achieved when the target handheld object is accurately placed at the target position, as evaluated using APIs provided by the world knowledge manager. In all benchmarks, the agent must execute the \texttt{STOP} action within a life horizon of 14,400 physical simulation steps; failure to do so results in the episode being deemed unsuccessful.

\noindent\textbf{Object Loco-Navigation Metrics.} Following prior navigation benchmarks~\cite{savva2019habitat, room2room, Matterport3D}, we employ four metrics: 1) \textbf{SR}: success rate, the \textit{primary metric}; 2) \textbf{PL}: path length, indicating the distance traveled by the agent; 3) \textbf{SPL}: success rate weighted by normalized inverse path length~\cite{anderson2018evaluation}, balancing success rate against trajectory length; and 4) \textbf{RT}: reset times, indicating the number of times the agent has fallen and been reset to the standing pose at the current position.\footnote{Because the current locomotion policy of H1 cannot enable it to crawl up by itself, we set the robot to a standing pose when it falls down, allowing it to continue the task.}

\noindent\textbf{Social Loco-Navigation Metrics.} In addition to SR, PL, SPL, and RT, we introduce \textbf{ECR} (excluded candidate rate) for Benchmark 2. Initially, the candidates are filtered by category name in the initial candidate list. After each query from the agent, the object candidates in the temporary list are filtered, and the ratio between the remaining candidates and the initial objects is computed to obtain ECR.

\noindent\textbf{Loco-Manipulation Metrics.} In addition to SR, PL, and RT, we introduce \textbf{SCR} (satisfied condition rate) for Benchmark 3, which measures the ratio of satisfied conditions within all conditions.

\begin{table}
\centering
\renewcommand\arraystretch{1.3}
\caption{\small
Quantitative results in GRBench using VLM baselines and LLM agent baselines. Metrics include: PL (Path Length), SR (Success Rate), SPL (Success rate weighted by normalized inverse Path Length), ECR (Excluded candidate rate), SCR (Satisfied condition rate), and RT (Reset times).
} % \caption
\vspace{1ex}
\resizebox{\linewidth}{!}{ %< auto-adjusts font size to fill line
\begin{tabular}{@{}|c|cc|cccc|ccccc|cccc|@{}}
\hline
\multicolumn{1}{|c|}{\multirow{2}*{Split}} & \multicolumn{2}{c|}{\multirow{2}*{Method}} & \multicolumn{4}{c|}{Object Loco-Navigation} & \multicolumn{5}{c|}{Social Loco-Navigation} & \multicolumn{4}{c|}{Loco-Manipulation}  \\
\cline{4-16}
~ & ~ & ~ & PL & SR & SPL & RT  & PL & SR & SPL & ECR & RT  & PL & SR & SCR & RT \\
\hline
% \multicolumn{2}{|c|}{Human} & \\
% \hline
\multirow{9}*{\rotatebox{90}{\texttt{validation} }} & \multicolumn{2}{c|}{Random} & 19.78 & 2.50 & 1.42 & - & 20.21 & 3.00 & 1.56 & - & - & 0.30 & 0 & 0 & - \\
% \hline
\cline{2-16}
~ & \multirow{3}*{\rotatebox{90}{VLM }}&\multicolumn{1}{|c|}{Qwen-VL~\cite{Qwen-VL}} & 11.43 & 8.00 & 4.30 & 0.18  & 0.60 & 1.00 & 0.78 & 0.92 & 0.28  & 0.11 & 0 & 0 & - \\
~ & ~ &\multicolumn{1}{|c|}{InternVL-Chat-1.5~\cite{chen2024far}} & 12.77 & 8.00 & 5.45 & 0.67 & {15.19} & {0.00} &{ 0.00} & {0.00} & {6.67}  & 0.10 & 0 & 0 & - \\
~ & ~ &\multicolumn{1}{|c|}{GPT-4o~\cite{gpt4o}} & 11.93 & 7.00 & 3.28 & 0.17 & 6.55 & 6.00 & 5.13 & 1.00 & 0.96  & 0.10 & 0 & 0 & - \\
\cline{2-16}
~ & \multirow{5}*{\rotatebox{90}{LLM }}&\multicolumn{1}{|c|}{Qwen~\cite{bai2023qwen}} & 25.27 & 19.00 & 12.86 & 0.67  & 20.72 & 14.00 & 9.33 & 17.79 & 1.66 & 0.23 & 0 & 0 & - \\
~ & ~&\multicolumn{1}{|c|}{Llama-3 8B~\cite{llama3}} & 25.97 & 10.00 & 7.21 & 0.84 & 20.77 & 5.00 & 3.49 & 17.69 & 1.54  & 0.21 & 0 & 0 & - \\
~ & ~ &\multicolumn{1}{|c|}{InternLM-2-Chat~\cite{cai2024internlm2}} & 22.29 & 18.00 & 10.97 & 0.42 & 22.25 & 11.00 & 6.37 & 4.93 & 1.79 & 0.23 & 0 & 0 & - \\
~& ~ &\multicolumn{1}{|c|}{ChatGLM3~\cite{zeng2022glm}} & 22.75 & 22.00 & 12.97 & 0.60  & 21.13 & 8.00 & 4.61 & 7.1 & 2.06 & 0.26 & 0 & 0 & -  \\
~ & ~ &\multicolumn{1}{|c|}{GPT-4o~\cite{gpt4o}}  & 23.36 & 10.00 & 5.87 & 0.87 & 22.27 & 7.00 & 4.49 & 1.02 & 2.82  & 0.25 & 0 & 0 & - \\

\hline

\multirow{8}*{\rotatebox{90}{\texttt{test} }} & \multirow{3}*{\rotatebox{90}{VLM }}&\multicolumn{1}{|c|}{Qwen-VL~\cite{Qwen-VL}} & 12.76 & 8.00 & 6.07 & 0.27 & 1.30 & 0.00 & 0.00 & 0.69 & 0.38 & 0.15 & 0 & 0 & -\\
~ & ~ &\multicolumn{1}{|c|}{InternVL-Chat-1.5~\cite{chen2024far}} & {10.85} & {5.50} & {3.88} & {0.23}& {16.80} & {0.00} & {0.00} & {0.00} & {6.35} & 0.13 & 0 & 0 & -\\
~ & ~ &\multicolumn{1}{|c|}{GPT-4o~\cite{gpt4o}}  & 12.18 & 14.00 & 9.13 & 0.27 & 11.10 & 2.00 & 1.69 & 0.98 & 3.10 & 0.15 & 0 & 0 & - \\
\cline{2-16}
~ & \multirow{5}*{\rotatebox{90}{LLM }}&\multicolumn{1}{|c|}{Qwen~\cite{bai2023qwen}}  & 23.68 & 16.00 & 10.03 & 0.61 & 21.14 & 12.50 & 9.18 & 5.21 & 1.76  & 0.30 & 0 & 0 & -\\
~&~&\multicolumn{1}{|c|}{Llama-3 8B~\cite{llama3}} & 23.32 & 15.50 & 10.96 & 0.57 & 21.42 & 11.00 & 6.26 & 7.21 & 1.89 & 0.29 & 0 & 0 & -\\
~&~ &\multicolumn{1}{|c|}{InternLM-2-Chat~\cite{cai2024internlm2}}  & 22.62 & 21.50 & 12.45 & 0.62 & 20.97 & 7.50 & 4.40 & 6.32 & 2.22  & 0.33 & 0 & 0 & -\\
~&~ &\multicolumn{1}{|c|}{ChatGLM3~\cite{zeng2022glm}}  & 23.78 & 22.00 & 11.68 & 0.74  & 20.36 & 11.00 & 6.97 & 0.13 & 2.25 & 0.26 & 0 & 0 & - \\
~&~ &\multicolumn{1}{|c|}{GPT-4o~\cite{gpt4o}}   & 23.16 & 16.00 & 5.33 & 0.80 & 20.88 & 7.00 & 5.10 & 5.39 & 2.16 & 0.27 & 0 & 0 & -\\

\hline

\end{tabular}
} % \resizebox
% \vspace{-3ex}
\label{tab:benchmark}
\end{table}
% \section{Experiments}
% \label{sec:experiments}
\vspace{-1.5ex}
\subsection{Quantitative Results}
\label{sec:experiments}
\vspace{-1.5ex}
In this section, we present a comparative analysis of our large model-driven agent framework under different large model backends across three benchmarks. As shown in Tab.~\ref{tab:benchmark}, we observe that a random strategy yielded a performance close to 0, indicating that our task is non-trivial. When utilizing a relatively superior large model as the backend, we achieve significantly better overall performance, consistently across all three benchmarks. Specifically, we observed that Qwen outperformed GPT-4o in dialogue.

Moreover, our agent framework demonstrates markedly superior performance compared to directly employing a multimodal large model for decision-making. This suggests that even the most advanced multimodal large models currently lack strong generalization capabilities in real-world embodied tasks. However, our approach also has considerable room for improvement. This indicates that when closer-to-real-world task settings are introduced, even tasks like navigation, which have been studied for many years, remain far from being fully solved.

For the loco-manipulation task, we observe a success rate (SR) of 0. Upon analysis, we identified two primary failure causes. First, failures during the locomotion process resulted in a significantly higher number of resets compared to object loco-navigation and social loco-navigation, where the reset counts were similar. We hypothesize that the main reason for this is the larger turning radius and base size of Aliengo compared to H1, making collisions more likely, thus highlighting the flexibility advantage of humanoid robots. Second, during manipulation, the arm frequently collides with other objects in the environment, indicating that current multimodal large models struggle with such complex motion planning. 

Finally, the result shows that the difficulty of these three tasks is incremental, providing a good gradation of challenge for evaluation. It is consistent with our intuitions behind task designs: 
(a) task-oriented scene-aware conversation is more challenging compared to instruction understanding, and (b) the loco-manipulation task is much more difficult than both navigation and static manipulation tasks due to the larger action space, longer planning horizon, and the requirement of more precise navigation and whole body collaboration.

\vspace{-1.5ex}
\subsection{Diagnostic Study}
\vspace{-1.5ex}
\begin{table}
\centering
\renewcommand\arraystretch{1.1}
\caption{\small
Diagnostic study on the components of our agent in 50 episodes from the \texttt{validation} split.
} % \caption
\vspace{0.5ex}
\resizebox{\linewidth}{!}{ %< auto-adjusts font size to fill line
\begin{tabular}{@{}|c|c|cccc|ccccc|cccc|@{}}
\hline
\multirow{2}*{\#} & \multicolumn{1}{c|}{\multirow{2}*{Perception}} & \multicolumn{4}{c|}{Object Loco-Navigation} & \multicolumn{5}{c|}{Social Loco-Navigation} & \multicolumn{4}{c|}{Loco-Manipulation}  \\
\cline{3-15}
~ & ~ & PL & SR & SPL & RT & PL & SR & SPL & EC &RT  & PL & SR & SC & RT\\
\hline
1 & GPT-4o~\cite{gpt4o} & 23.82 & 8.00 & 4.26 & 0.89 & 21.70 & 12.00 & 9.88 & 1.57 & 1.80 & 0.23 & 0 & 0 & -  \\
2 & Qwen-VL~\cite{Qwen-VL} & 24.67 & 6.00 & 4.17 & 0.87 & 21.41 & 12.00 & 10.36 & 0.56 & 2.30 & 0.23 & 0 & 0 & - \\
% \hline
% 3 & GPT-4o & Teleportation  \\
\hline
\end{tabular}
} % \resizebox
% \vspace{-3ex}
\label{tab:abla_component}
\end{table}
In this section, we present the results of ablation studies conducted to verify the impact of various factors on model performance and experimental outcomes. These studies were aimed at understanding the influence of task settings and model components on the performance of our agent framework.

First, we compare the effects of the perception module on the experimental results. As shown in Tab.~\ref{tab:abla_component}, we replace the GPT-4o, responsible for the perception part of the agent framework, with QwenVL and conducted an evaluation. The results indicate that perception performance has a significant impact on overall performance. Specifically, the use of QwenVL leads to superior results compared to GPT-4o. 
We tested the use of an RL controller versus an oracle action setting where positions are directly set. The oracle actions significantly enhanced performance, validating our hypothesis that there is a substantial gap between this oracle action assumption and real-world application scenarios. This gap indicates that while oracle actions provide an idealized performance, real-world implementations must contend with more complex and less predictable movement dynamics.

\subsection{Qualitative Results}
\begin{figure}
    \centering
    \vspace{-15pt}
    \includegraphics[width=0.90\linewidth]{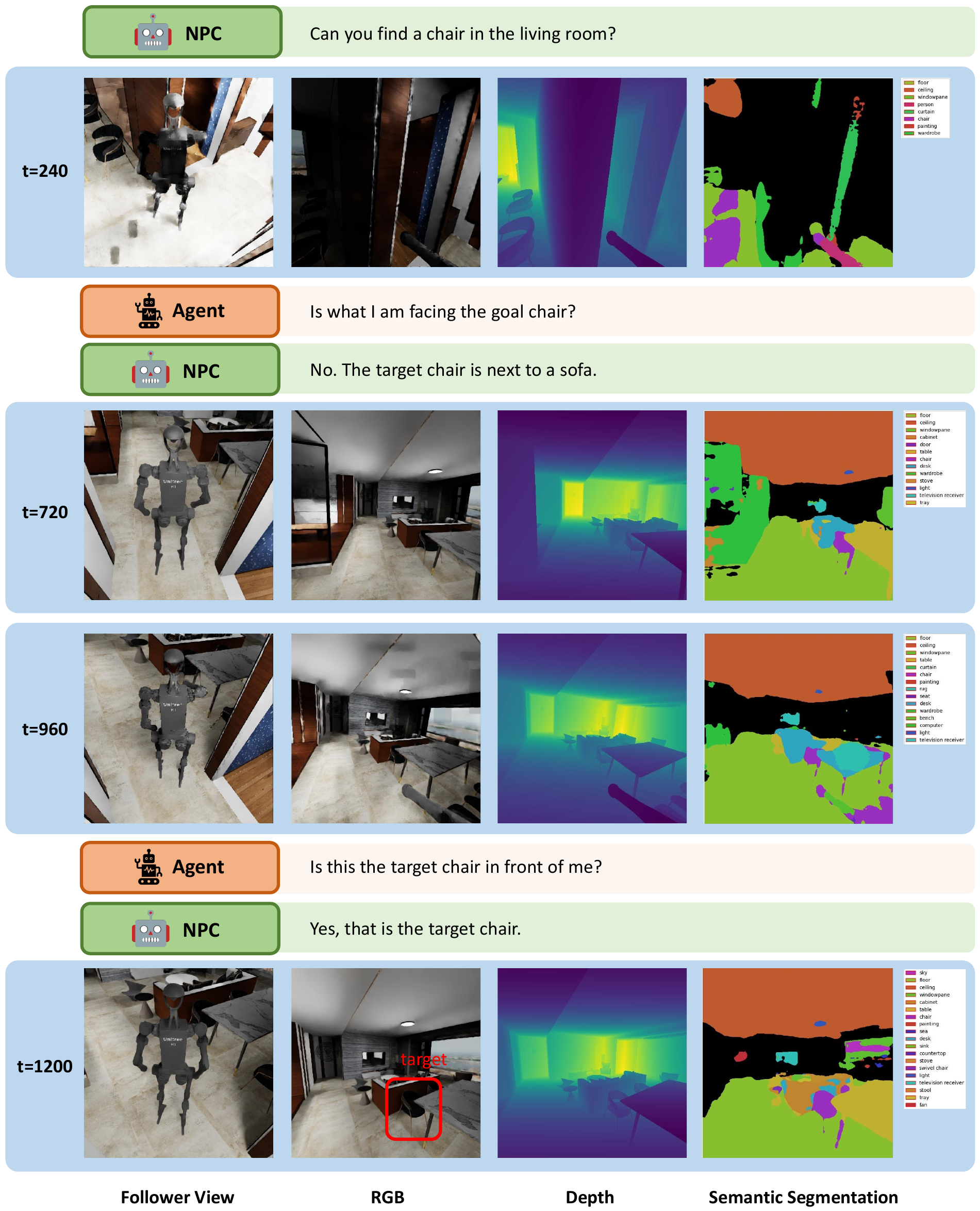}
    \caption{The agent successfully finds the target object with the help of the NPC.}
    \vspace{-10pt}
    \label{fig:case}
\end{figure}

 We showcase an episode performed by our LLM agent on the Social Loco-Navigation task in Fig.~\ref{fig:case} to illustrate how the agent interacts with the NPC. The agent is allowed to talk to the NPC up to three times to query additional task information. At $t=240$, the agent navigates to a chair and asks the NPC whether it is the target chair. The NPC then provides surrounding information about the goal to mitigate ambiguity.  With the NPC's assistance, the agent successfully identifies the target chair through an interactive process akin to human behavior. This demonstrates that our NPC is capable of providing natural social interactions for studying human-robot interaction and collaboration.

\vspace{-2ex}
\section{Conclusions}
\vspace{-1.5ex}
\label{sec:conclusions}
In this paper, we present GRUtopia, a novel platform designed to stimulate and benchmark advanced embodied AI research. GRUtopia offers a city-scale environment with diverse functional scenarios (GRScenes), learning-based control APIs for various embodiments, and multi-modal NPCs capable of social interaction, task generation, and task assignment. Utilizing our scene dataset and utilities, we propose GRBench, which currently includes three benchmarks to assess robots' capabilities in navigation, social interaction, manipulation. We conducted extensive experiments to thoroughly analyze our benchmark and the performance of state-of-the-art large vision-language models (VLMs). GRUtopia is in active development. In the initial version, we partially release 100 annotated, ready-to-use indoor scenes and a city block. The current NPC system supports social interaction without physically realistic contact. We will continuously enhance the platform's features, including 3D scene assets, control policies, task generation, NPC systems, and benchmarks, to facilitate the scaling up of embodied learning.

\textbf{Acknowledgement.} We clarify author contributions as follows: Jiangmiao Pang initiated and led the project. Hanqing Wang directed the project development. Tai Wang, Jiahe Chen, and Peizhou Cao contributed to GRScenes. Boyu Mi focused on GRResidents. Jiahe Chen, Wensi Huang, Siheng Zhao, and Yilun Chen contributed to GRBench. Qingwei Ben, Zirui Wang, and Junfeng Long trained locomotion policies. Tao Huang and Wenye Yu were responsible for manipulation policies. Sizhe Yang provided various utilities. Zichao Ye, Jialun Li, Ying Zhao, and Zhongying Tu were responsible for platform development. Huiling Wang served as the project manager. Yu Qiao and Dahua Lin provided project advisement.
%%%%%%%%% REFERENCES
{
    % \clearpage
    \small
    \bibliographystyle{ieee_fullname}
    \bibliography{macros,main}
}
% --- supplementary material
\clearpage
\appendix

\renewcommand\thesection{\Alph{section}}
\renewcommand\thefigure{\Alph{figure}}
\renewcommand\theequation{\Alph{equation}}
\renewcommand\thealgorithm{\Alph{algorithm}}
\renewcommand\thetable{\Alph{table}}
\renewcommand\thepage{S\arabic{page}}

\section{More Details of GRScenes}
\begin{figure}
\caption{Two examples of meta annotations in GRScenes (Sec.~\ref{sec:grscenes}). The example objects are outlined in orange. Notably, the part labels and material labels are embedded within the object assets.}
\vspace{5pt}
% \centering
\begin{minipage}[t]{0.45\textwidth}
    \begin{subfigure}[t]{\textwidth}
         \centering
         \includegraphics[width=\textwidth]{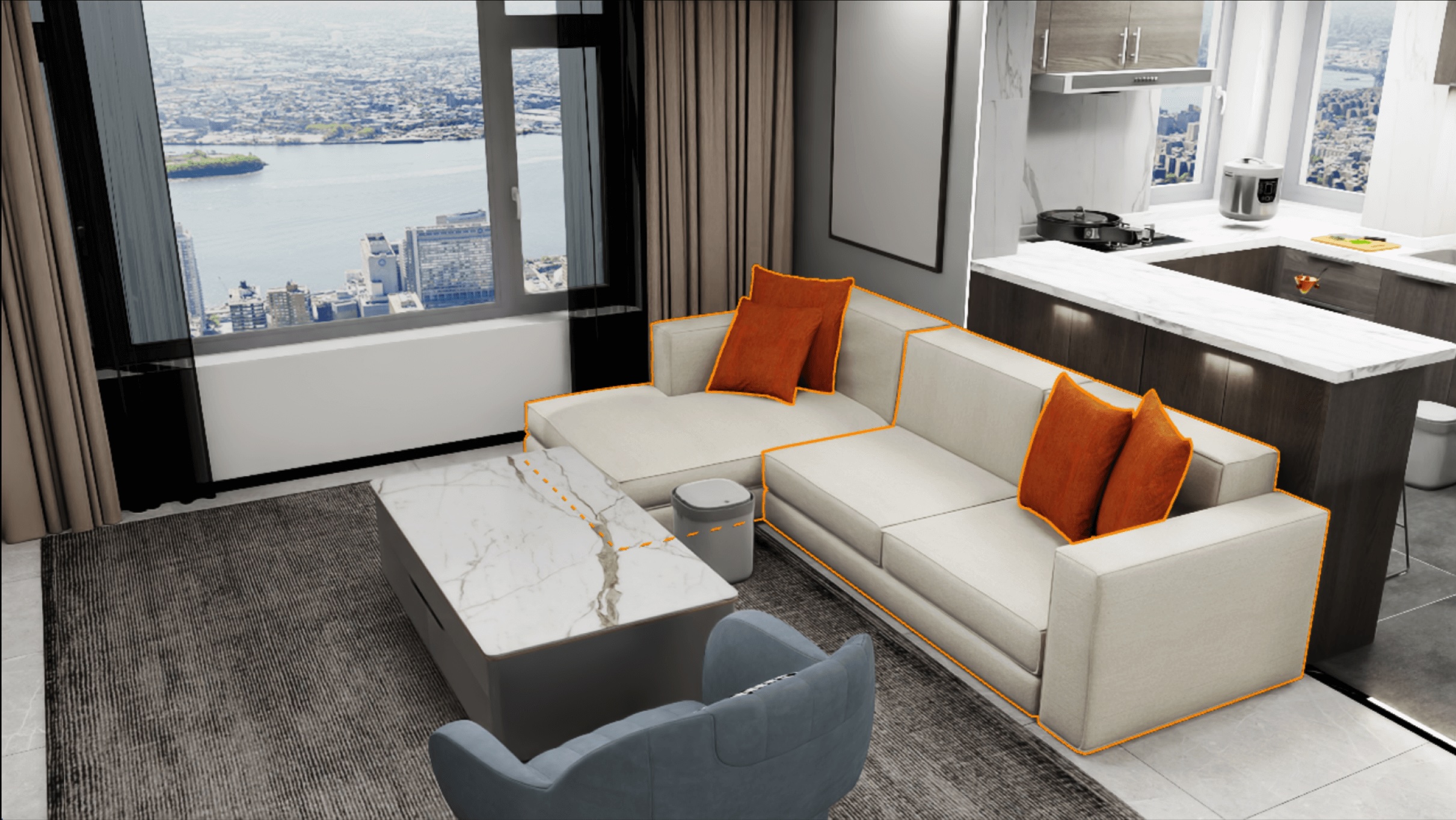}
    \end{subfigure}
    \begin{subfigure}[b]{0.15\textwidth}
         \centering
         \includegraphics[width=\textwidth]{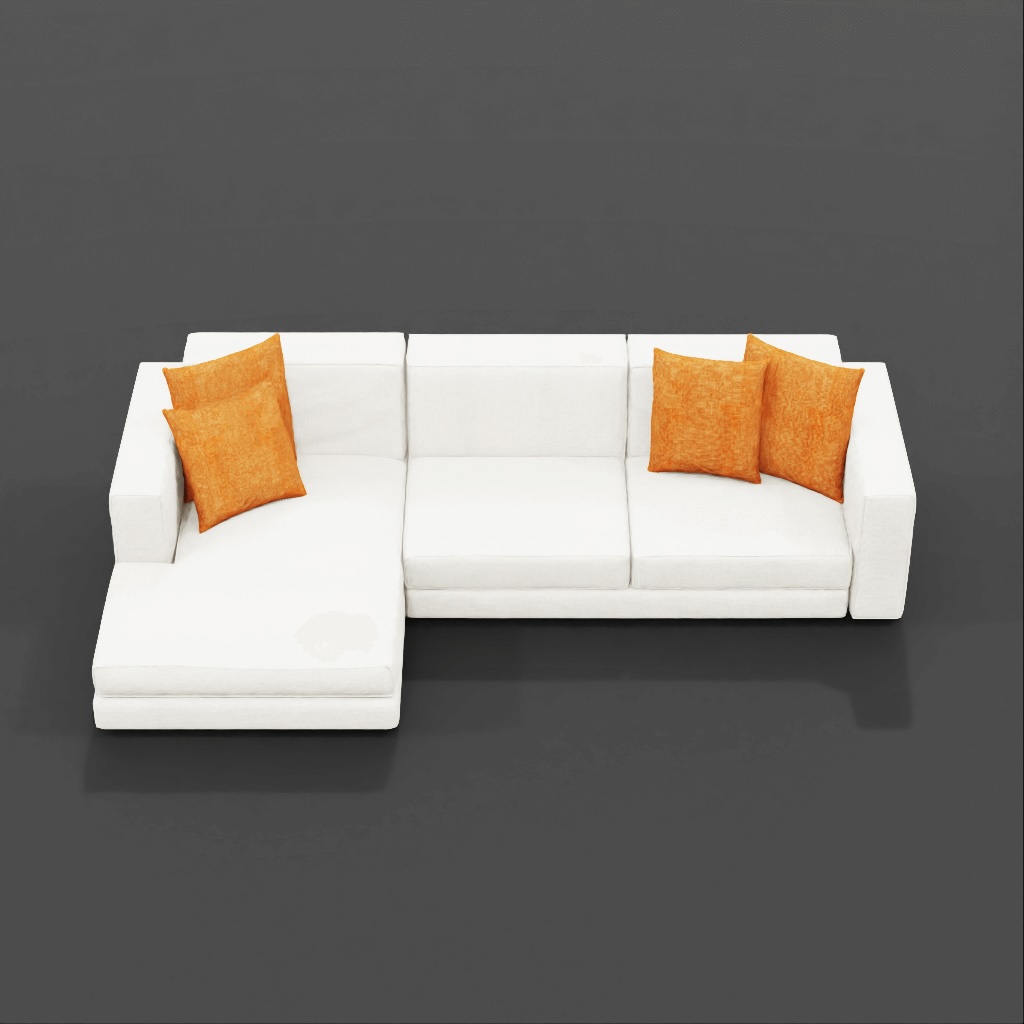}
         \caption*{\tiny View 1}
    \end{subfigure}\hfill
    \begin{subfigure}[b]{0.15\textwidth}
         \centering
         \includegraphics[width=\textwidth]{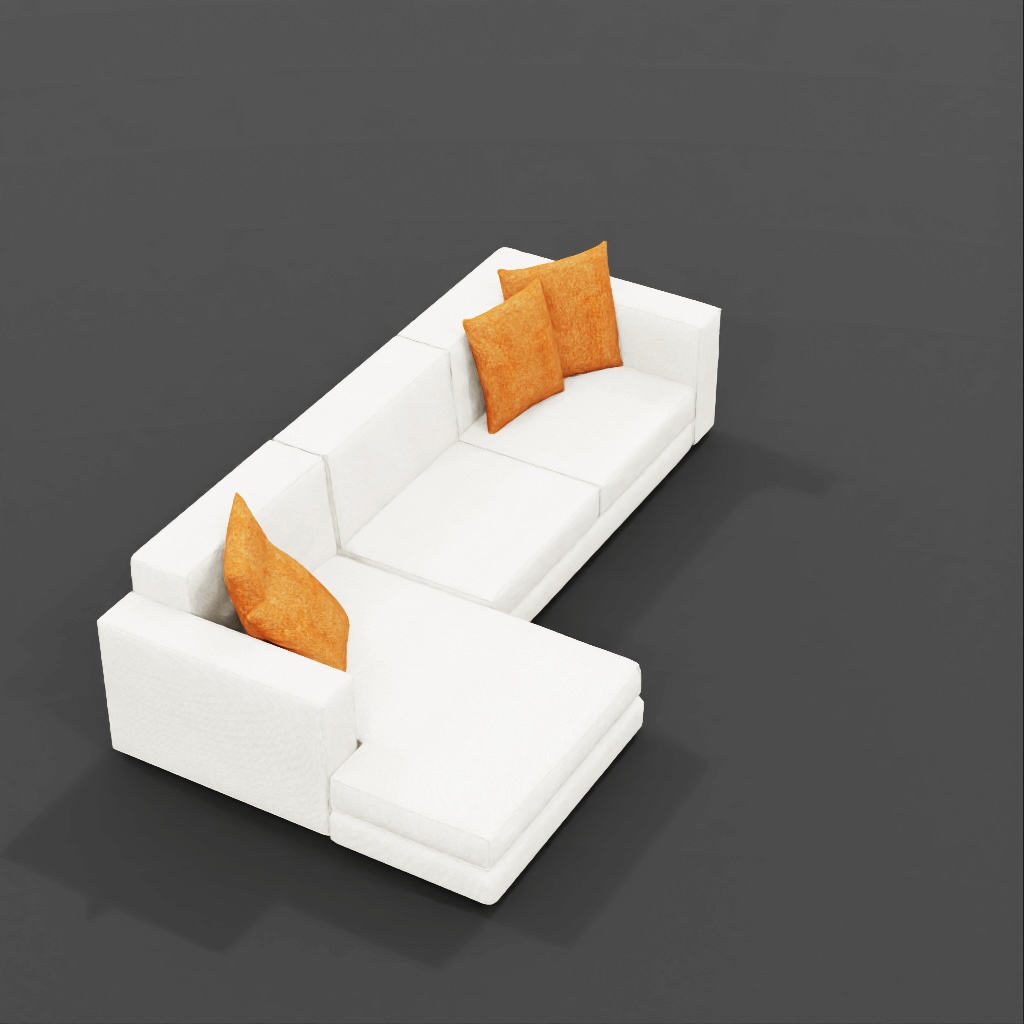}
         \caption*{\tiny View 2}
    \end{subfigure}\hfill
    \begin{subfigure}[b]{0.15\textwidth}
         \centering
         \includegraphics[width=\textwidth]{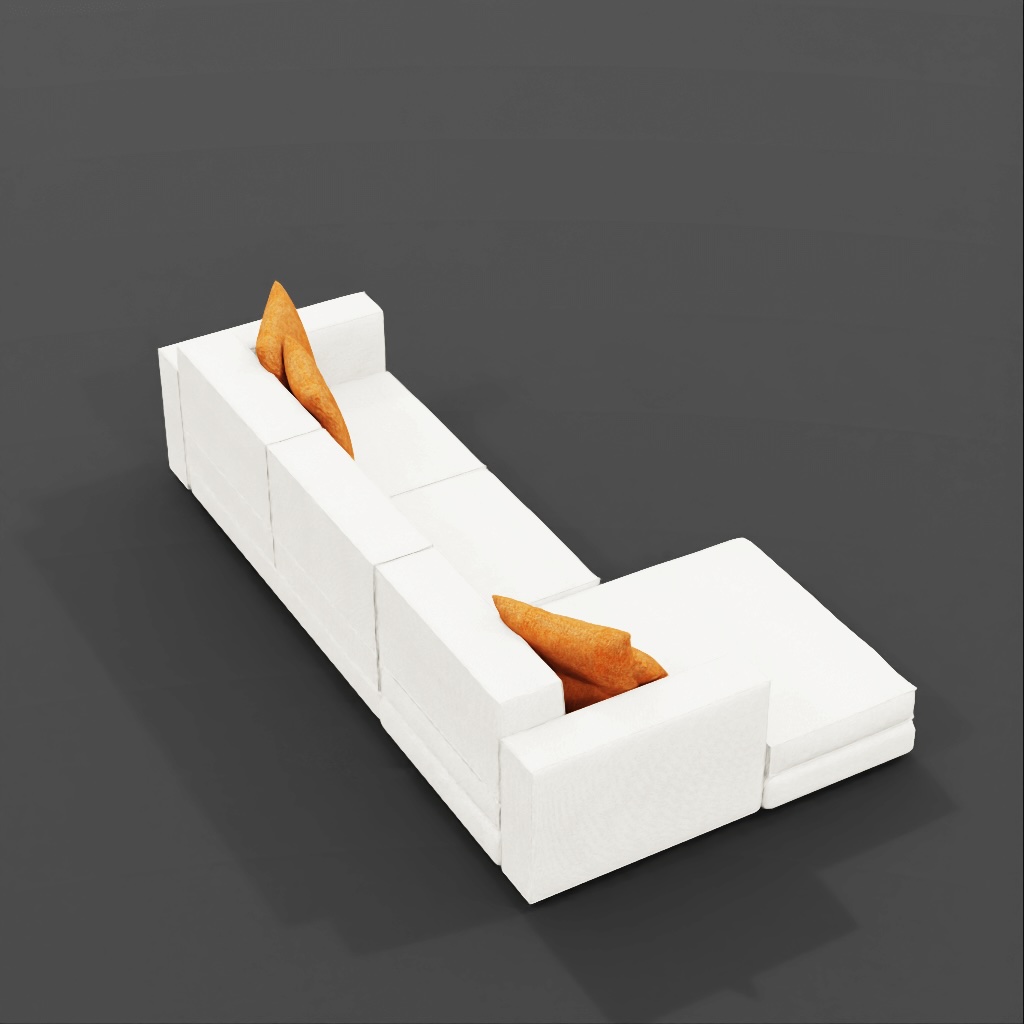}
         \caption*{\tiny View 3}
    \end{subfigure}\hfill
    \begin{subfigure}[b]{0.15\textwidth}
         \centering
         \includegraphics[width=\textwidth]{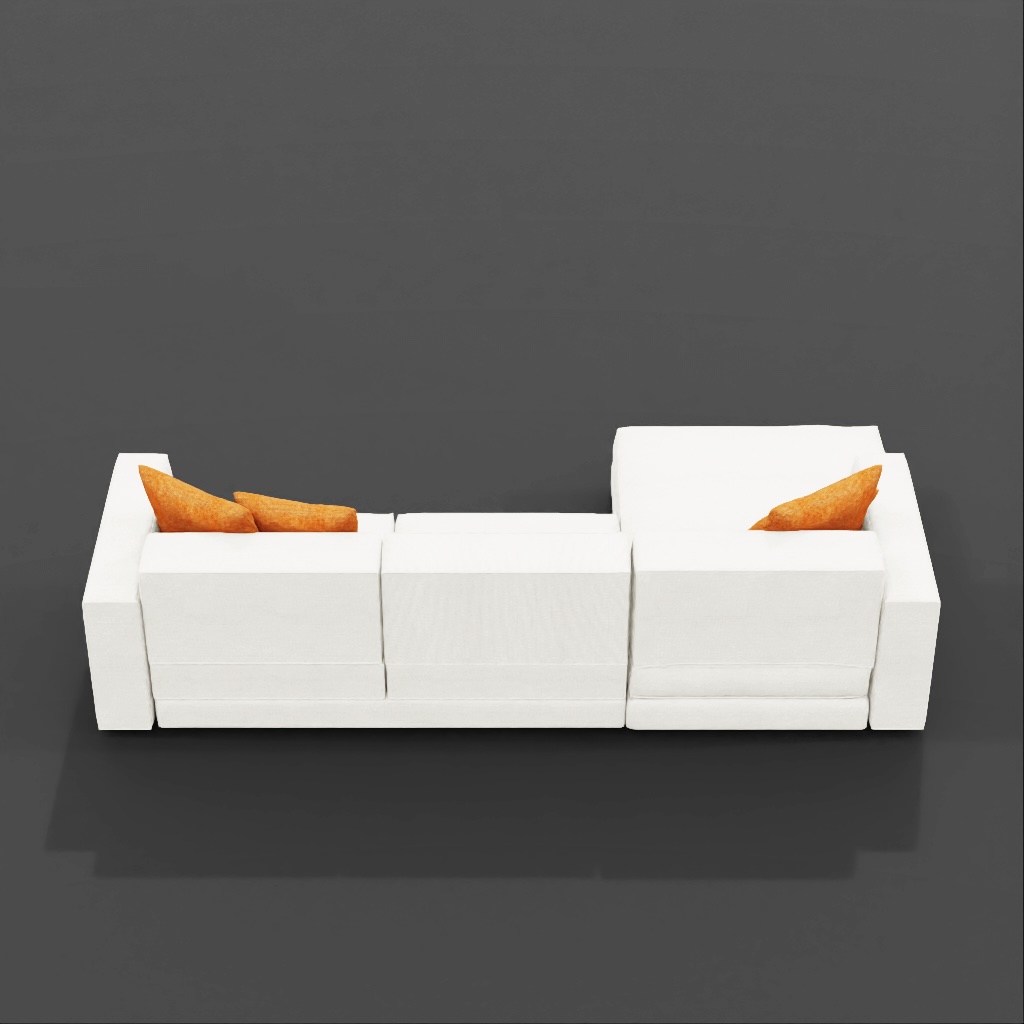}
         \caption*{\tiny View 4}
    \end{subfigure}\hfill
    \begin{subfigure}[b]{0.15\textwidth}
         \centering
         \includegraphics[width=\textwidth]{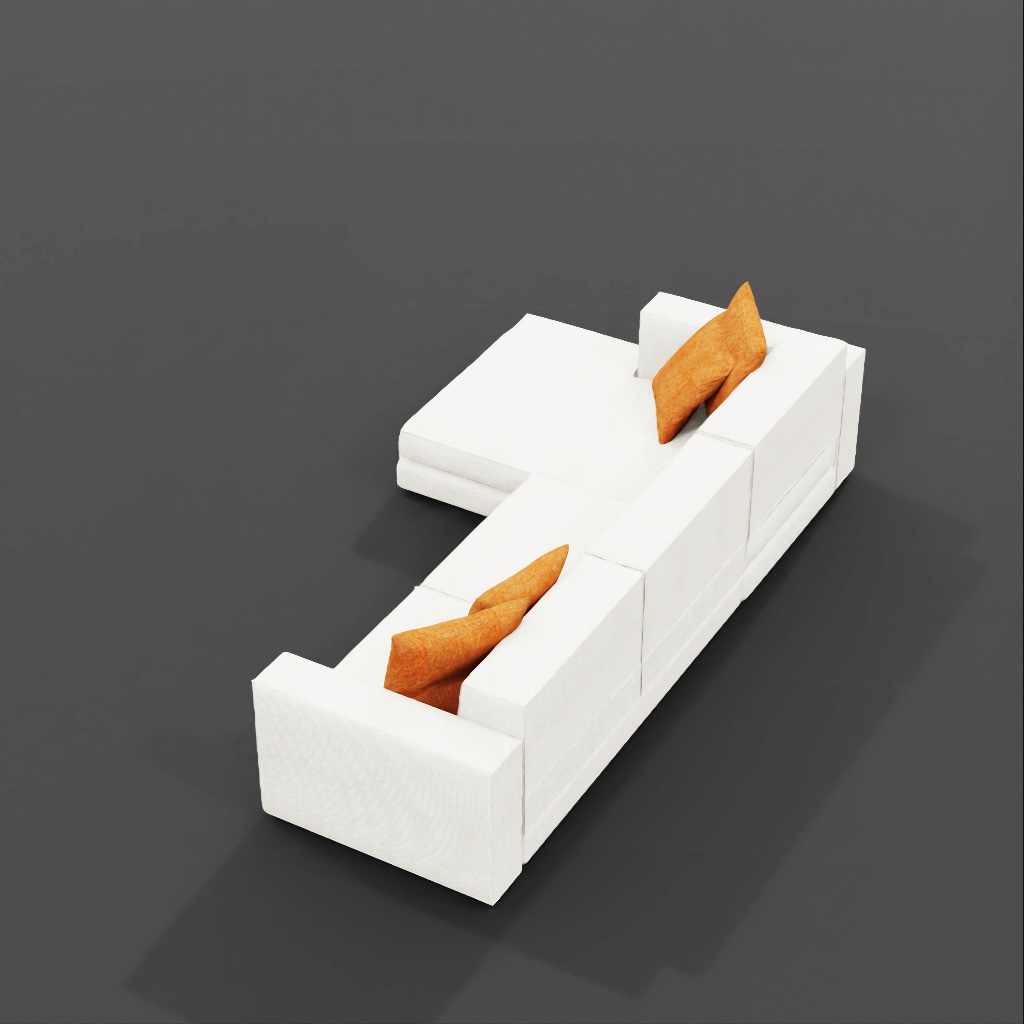}
         \caption*{\tiny View 5}
    \end{subfigure}\hfill
    \begin{subfigure}[b]{0.15\textwidth}
         \centering
         \includegraphics[width=\textwidth]{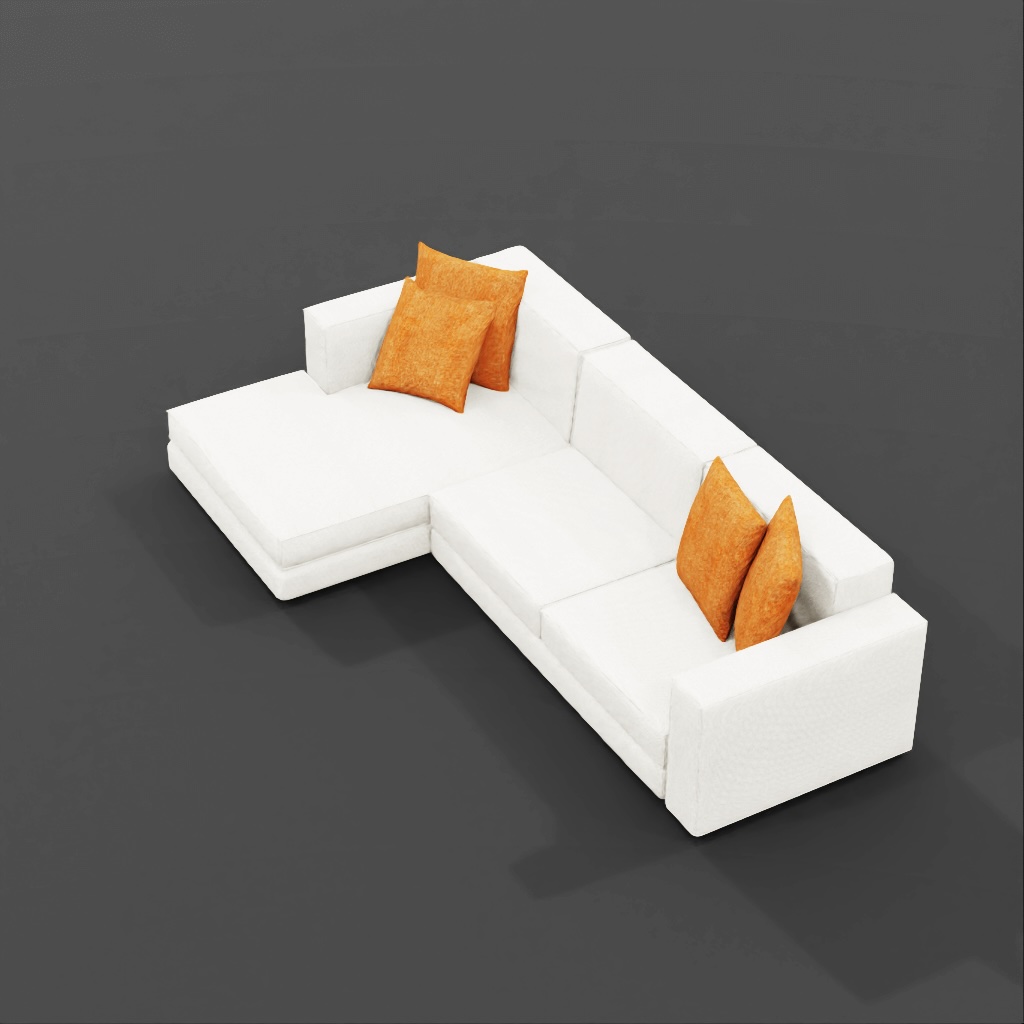}
         \caption*{\tiny View 6}
    \end{subfigure}
    \begin{minted}[fontsize=\tiny]{json}
{
    "couch/SM_01_6D7YPDMTDD522XTVKQ888888": {
    "instance_id": "couch/SM_01_6D7YPDMTDD522XTVKQ888888",
    "category": "couch",
    "scope": "Furnitures",
    "room": "1/living room",
    "position": [
        2.008520397463428,
        1.5190480830478437,
        0.408894387374766
    ],
    "min_points": [
        1.256558941300776,
        0.09561449997743492,
        -7.080078034960025e-09
    ],
    "max_points": [
        2.7604818536260796,
        2.9424816661182525,
        0.8177887818296101
    ],
    "description": [
        "The object is a white L-shaped couch.",
        "It has orange cushions.",
        "The couch has a rectangular base.",
        "There is a chaise lounge on one end.",
        "A backrest extends along the length of the couch.",
        "The couch has visible seams.",
        "It is made of a textured fabric.",
        "The cushions are square.",
        "The cushions have a smooth texture."
    ],
    "nearby_objects": {
        "window/SM_05_628FFE90_19FA_4477_9708_F431EF1D3347": [
            "near",
            0.5333050583827639
        ],
        "light/SM_01_6IMBRKRVAVAACPTULY888888": [
            "near",
            2.0458988819787214
        ],
        "sofachair/SM_01_6PRLMUJVAJTTWPTUKM888888": [
            "near",
            0.8054710046610357
        ],
        "curtain/SM_09_6DJUMTMTD5TGAGWBJQ888888": [
            "near",
            0.14992408896487763
        ],
        "curtain/SM_10_6DJUMTMTD5TGAGWBJQ888888": [
            "near",
            0.245047940131119
        ],
        "blanket/SM_04_6MOOWWZVAUTV4PTUKI888888": [
            "under",
            0.02822649117670297
        ],
        "picture/SM_01_6PVLJ4JVAV6CAPTUJU888888": [
            "above",
            0.18700055798224122
        ],
        "trashcan/SM_04_ZFEFG2JVALHVSPTUJU888888": [
            "below",
            0.1733666386620497
        ],
        "teatable/SM_01_ZFEFO5ZVAIJG6PTULY888888": [
            "near",
            0.23229627379782494
        ]
    }
}
    \end{minted}

\end{minipage}\hfill
\begin{minipage}[t]{0.45\textwidth}
\centering
    \begin{subfigure}[t]{\textwidth}
         \centering
         \includegraphics[width=\textwidth]{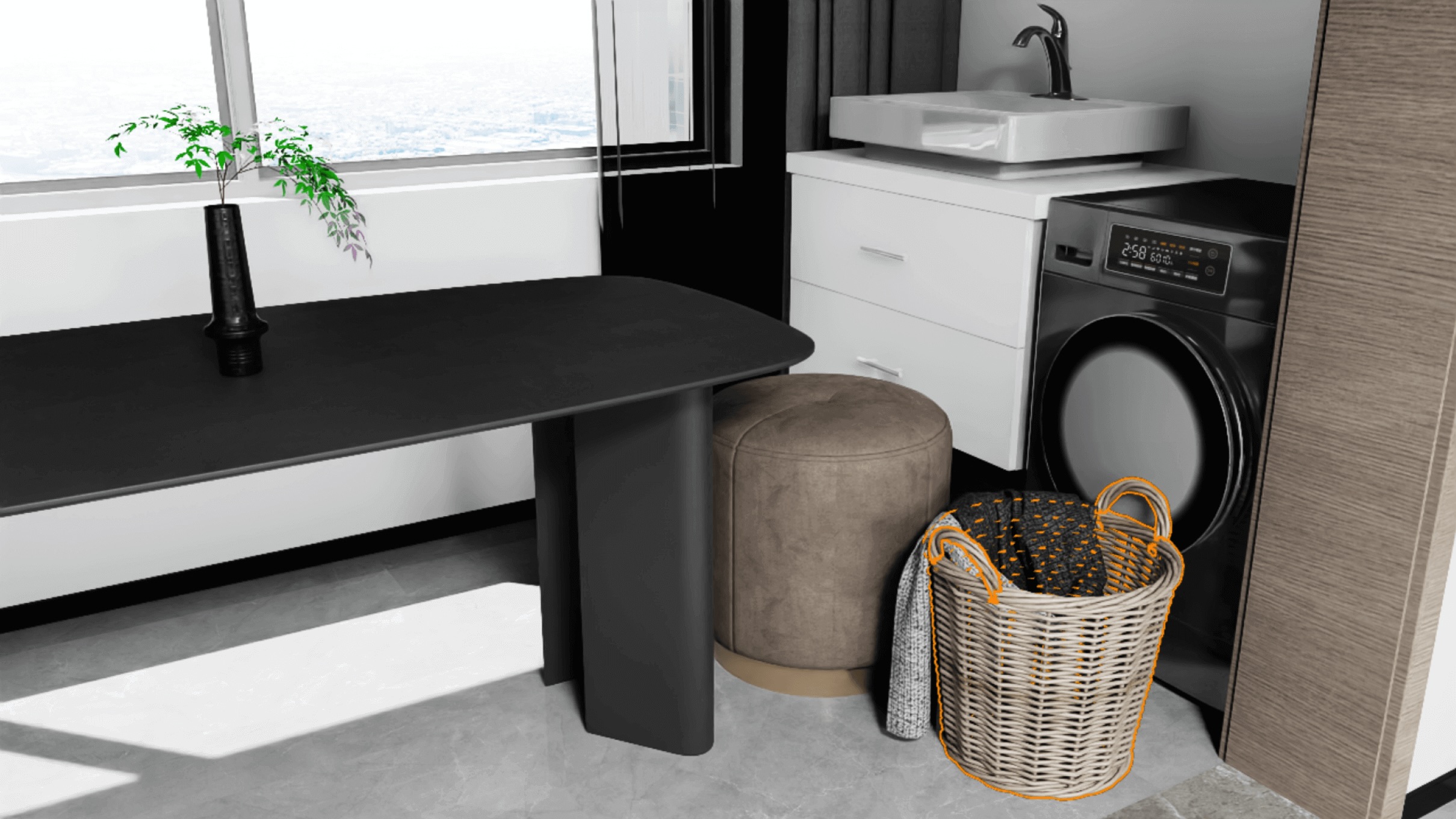}
    \end{subfigure}
    \begin{subfigure}[b]{0.15\textwidth}
         \centering
         \includegraphics[width=\textwidth]{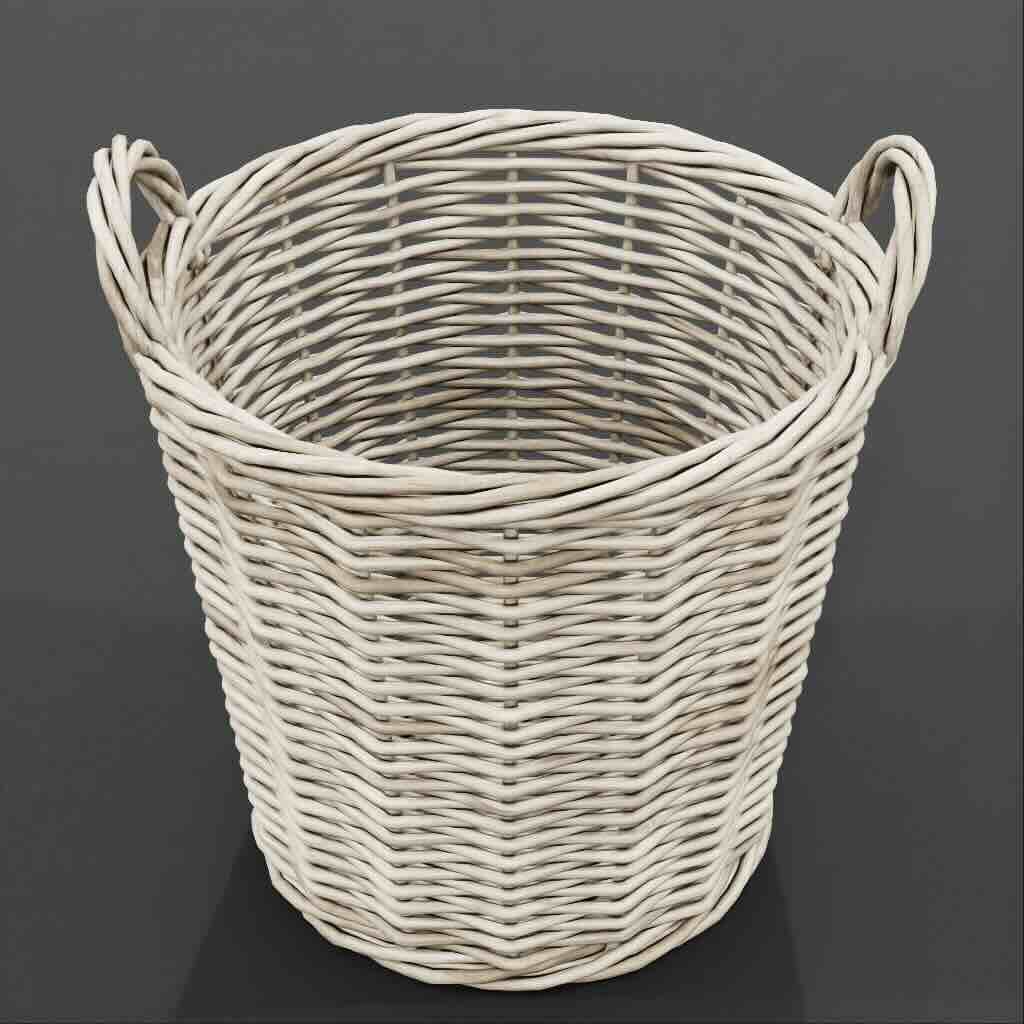}
         \caption*{\tiny View 1}
    \end{subfigure}\hfill
    \begin{subfigure}[b]{0.15\textwidth}
         \centering
         \includegraphics[width=\textwidth]{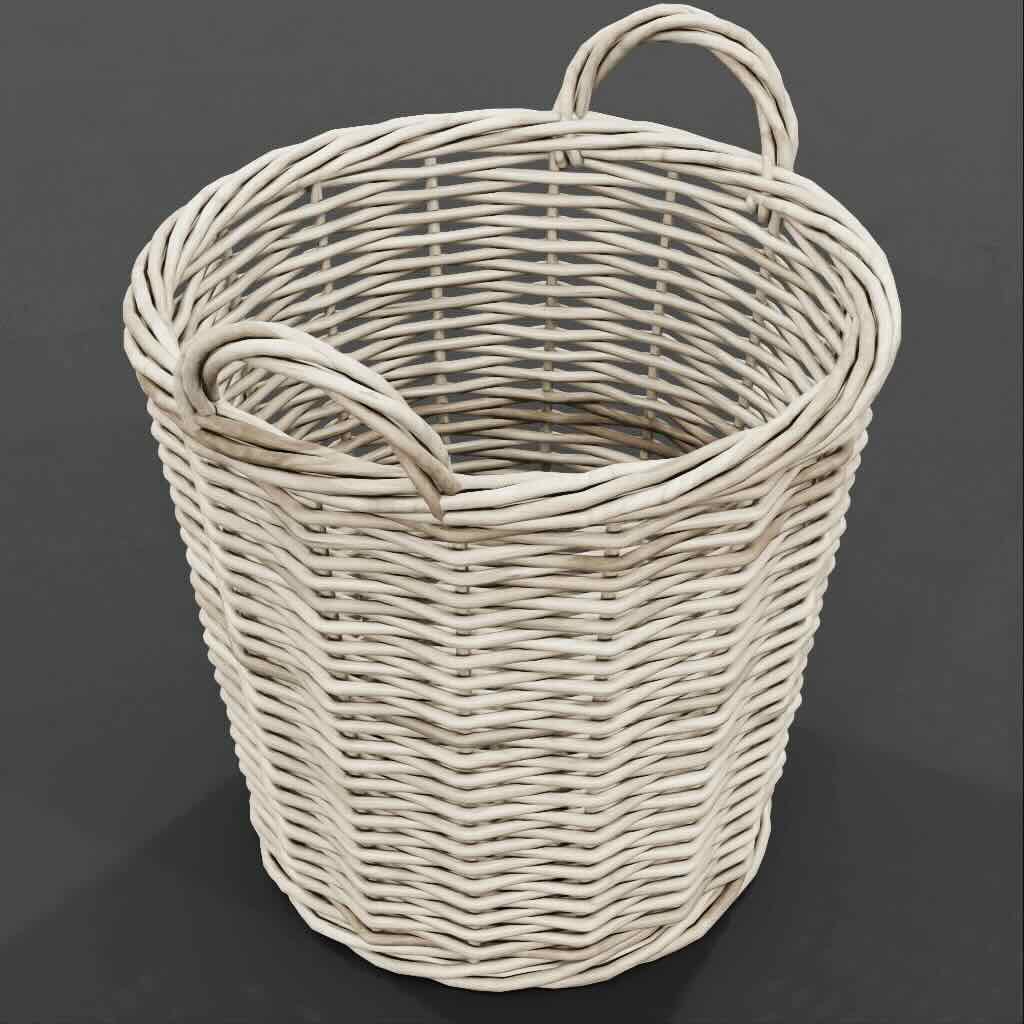}
         \caption*{\tiny View 2}
    \end{subfigure}\hfill
    \begin{subfigure}[b]{0.15\textwidth}
         \centering
         \includegraphics[width=\textwidth]{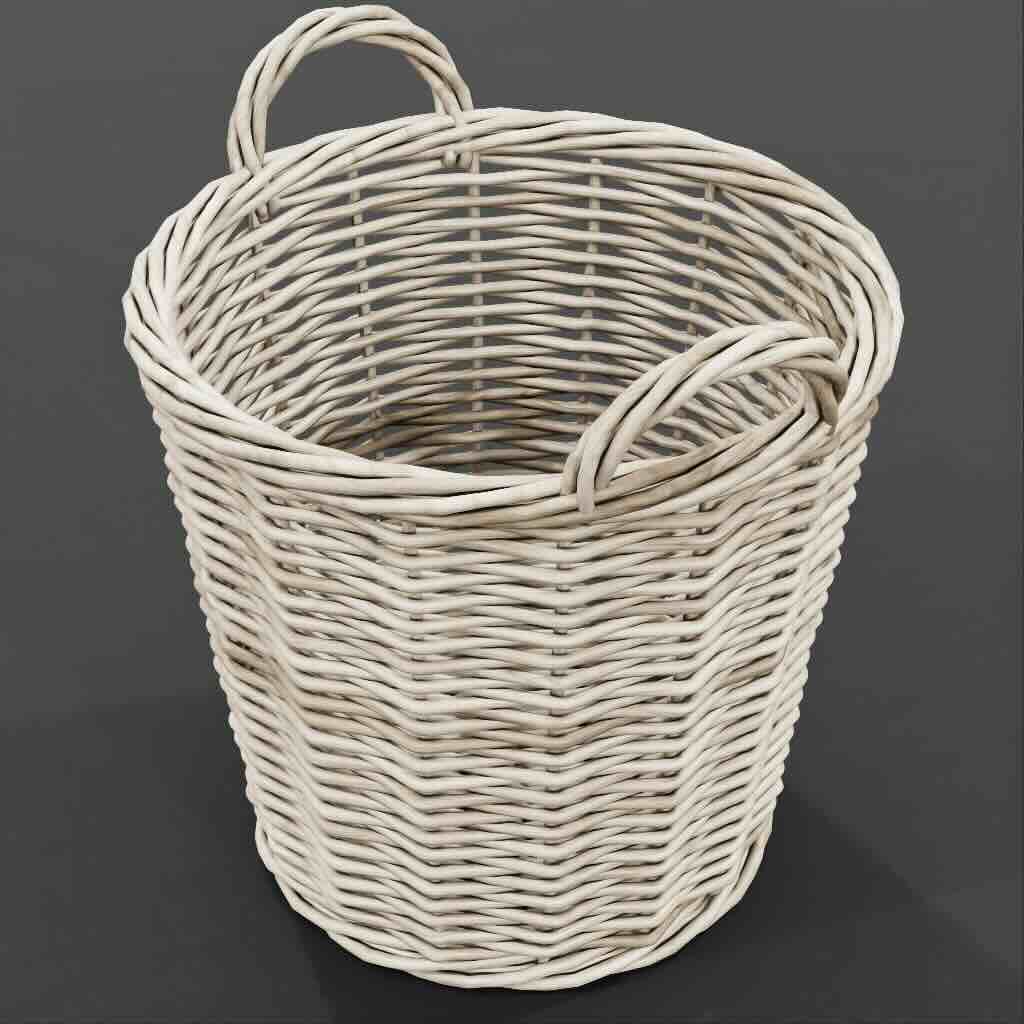}
         \caption*{\tiny View 3}
    \end{subfigure}\hfill
    \begin{subfigure}[b]{0.15\textwidth}
         \centering
         \includegraphics[width=\textwidth]{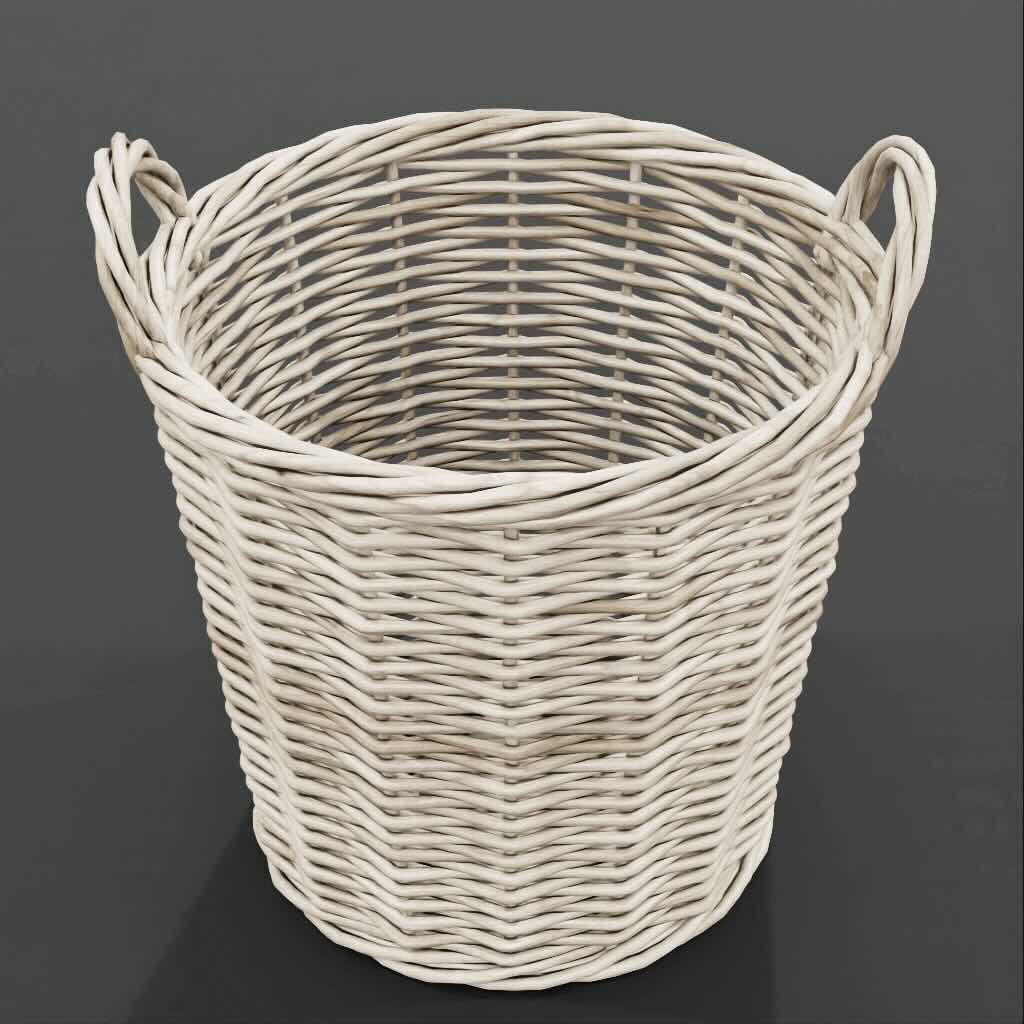}
         \caption*{\tiny View 4}
    \end{subfigure}\hfill
    \begin{subfigure}[b]{0.15\textwidth}
         \centering
         \includegraphics[width=\textwidth]{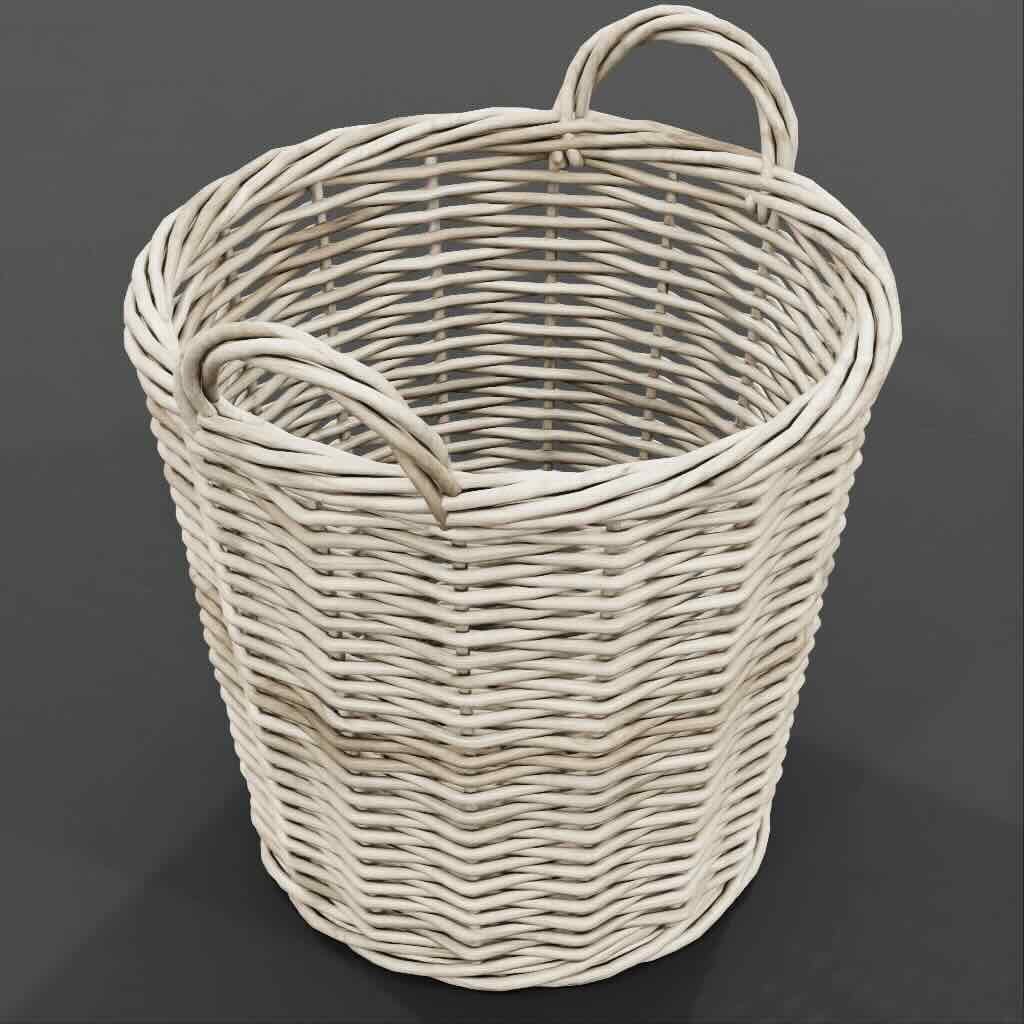}
         \caption*{\tiny View 5}
    \end{subfigure}\hfill
    \begin{subfigure}[b]{0.15\textwidth}
         \centering
         \includegraphics[width=\textwidth]{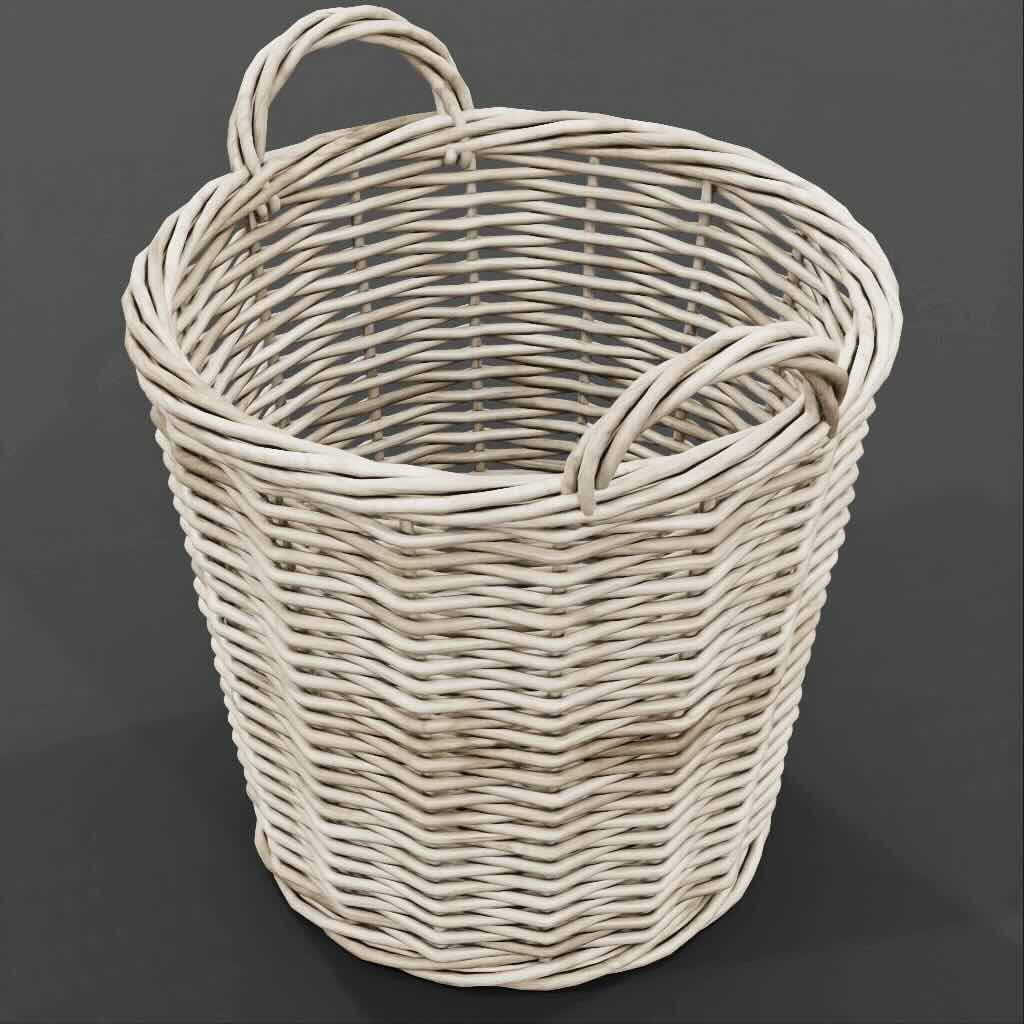}
         \caption*{\tiny View 6}
    \end{subfigure}
    \begin{minted}[fontsize=\tiny]{json}
{
    "basket/SM_00_6PY52XBVAV6OSPTUKQ888888": {
    "instance_id": "basket/SM_00_6PY52XBVAV6OSPTUKQ888888",
    "category": "basket",
    "scope": "Furnitures",
    "room": "7/balcony",
    "position": [
        -1.533837262909261,
        4.720350097208155,
        0.21521032797343898
    ],
    "min_points": [
        -1.7399600061728762,
        4.530086438764703,
        0.000541858890546278
    ],
    "max_points": [
        -1.3277145196456457,
        4.910613755651608,
        0.4298787970563317
    ],
    "description": [
        "The object is a basket with a cylindrical shape.",
        "It is made of woven material.",
        "The basket has a light beige color.",
        "The basket has two handles, one on each side.",
        "The handles are made of the same woven material.",
        "The weaving pattern is consistent throughout the basket.",
        "Vertical and horizontal strands create a grid-like structure.",
        "The top edge of the basket is reinforced with a thicker woven band."
    ],
    "nearby_objects": {
        "cabinet/SM_02_cabinet_1": [
            "near",
            0.3340383792697092
        ],
        "plant/SM_04_ZB2KC6ZVAJQQIPTUKU888888": [
            "near",
            0.9220090184554028
        ],
        "blanket/SM_04_6PY52XBVAV6OSPTUKQ888888": [
            "in",
            0.003548709679052394
        ],
        "blanket/SM_05_6PY52XBVAV6OSPTUKQ888888": [
            "in",
            0.004331730411871013
        ],
        "table/SM_04_ZB2KC6ZVAJQQIPTUKU888888": [
            "near",
            0.3451297420046326
        ],
        "sink/SM_01_6CK2X7MTDD52ANSSLE888888": [
            "near",
            0.6873307360030145
        ],
        "stool/SM_01_6GGTBW4TDD52APTUKA888888": [
            "near",
            0.07238074949219277
        ],
        "bottle/SM_00_ZB2KC6ZVAJQQIPTUKU888888": [
            "near",
            1.006464318234342
        ],
        "faucet/SM_04_ZEMERMZVAJSQIPTUKE888888": [
            "near",
            1.0355706964183513
        ],
        "washingmachine/SM_01_ZELJUSBVAJ5NEPTULM888888": [
            "near",
            0.0666639528382904
        ]
    }
}
    \end{minted}

\end{minipage}
\label{fig:meta_annotation}
\end{figure}

In this section, we provide more details of GRScenes with respect to 
data annotations (Fig.~\ref{fig:meta_annotation}) and scene graph extraction.

\noindent\textbf{Scene Graph Extraction.} We employ scene graphs \cite{kim20193} to structurally represent scenes, with each node describing an object with attributes such as position, size, and appearance information in textual form obtained by utilizing a visual language model on its multi-view images. Each edge within the scene graph denotes the relationships between objects. We follow \cite{referit3d} for relationship annotation. Notably, unlike scanned scenes that contain surface information only, the detailed internal structures in our scenes enable scene graphs to incorporate additional spatial relationships, namely \textit{in} and \textit{out of}. We extract color and shape information from object captions in the scene graph and identify the corresponding objects through exact text match, ensuring precise appearance-to-object alignment.

\section{More Details of GRResidents}

\subsection{Implementation Details of World Knowledge Manager}

Our world knowledge manager possesses three core functions (Sec.~\ref{sec: generative npc in the wild}) to support knowledge extraction from the environment. In this section, we provide Python-style pseudo-code in Algorithms~\ref{alg:diff}--\ref{alg:filter} to showcase their implementations.

\subsection{More Details of NPC Experiments}

\noindent\textbf{Object Grounding Experiment.}
We use GPT-4o\cite{gpt4o} to play the role of human annotator in object grounding experiment, see (Sec.~\ref{sec: generative npc in the wild}).

For the grounding utterance generation in the experiment, we randomly selected 10 home scenes and their corresponding scene graphs. Based on the appearance information from the object descriptions, we used the template \textit{<target object info> <relation> <anchor object info>} to automatically generate 500 object descriptions. Subsequently, we utilized GPT-4o to introduce a series of rule-based modifications, making the automatically generated template descriptions more challenging and natural. These modifications include:

\textit{Hiding the target object's category.} In real scenarios, common object descriptions might omit the category of the target object, such as "find the object on the table." Hiding the target object's category can make the descriptions more realistic and naturally increase the difficulty of the grounding task.

\textit{Relationship replacement.} As the relationships in the template are derived directly from a limited set of scene graph relationships, replacing them with synonyms, such as substituting "near" with "beside" or "next to", enhances the naturalness.

\textit{Sentence adjustment.} Modifying the original sentences by adding phrases such as "find the ..." or "I want to get ..." at the beginning of the utterances. This type of modification makes the utterances more similar to natural instructions, enhancing the realism and complexity of the descriptions.

For object grounding, we introduce a new API compatible with our framework:

 \texttt{\funcc{grounding(}\argc{target\_object\_info}, \argc{anchor\_object\_info}, \argc{relation\_name}\funcc{)}}

% This API can determine the target object based on the target object information, anchor object information, and spatial relationship.
Parameter \texttt{anchor\_object\_info} contains text format description on category, color or shape of the anchor object and target object.

This API first finds out the most suitable object in the scene based on the anchor object information as the anchor object for grounding.
Then among all other objects nearby the anchor object annotated in the scene graph, it can find out the target object by matching the spatial relationship between the anchor object and target object and parameter \texttt{relation\_name}.
Finally, it finds out the target object by matching \texttt{target\_object\_info} in the same process as finding the anchor object.

In our grounding experiment, with API definitions and task guidance, the NPC reasons through and composes proper arguments for the API call to get the grounding result.

\noindent\textbf{Object-Centric QA Experiment.}
We use a data generation pipeline combining GPT-4o~\cite{gpt4o} generation with manual verification. 
%prompt
Through the data-generation pipeline, we obtained 489 episodes of object-centric navigation. Each episode includes a target object, multiple rounds of agent actions, and natural language interactions with the NPC, covering a total of 1,669 interaction turns. 

For evaluation, we use the cosine similarity of sentence embeddings from \textit{text-embedding-3-large}~\cite{text-embedding-3-large} as the similarity metric. Based on a manual review of a subset of the data, we empirically set 0.6 as the similarity threshold. Answers with a similarity score above this threshold are marked as correct and receive a score of 100, while those below the threshold receive a score of 0. The final result is the average score across all QA pairs.

% \begin{center}
%     \scalebox{0.75}{
    \begin{algorithm}
    \caption{This function finds one difference that can shrink the candidate set. We design a hierarchical searching priority that has ``category'' $>$ ``room'' $>$ ``spatial relationship'' $>$ ``appearance'' to fit human habits in describing objects.}
    \label{alg:diff}
    \begin{minted}[fontsize=\tiny]{python}
def find_diff(self, target: str, candidates: list[str]):
    """
        return: (diff_type, difference) Tuple[str, list]
    """
    candidates = [(obj_id, self.spatial_relations[obj_id]) # scene graph
                                                for obj_id in candidates]
    categories = set()
    rooms = set()
    relations = defaultdict(list) 

    current_relation = {}
    for (obj_id, relation) in candidates:
        cate = relation["category"]
        room = relation["room"]
        categories.add(cate)
        rooms.add(room)
        relation_sets = {"near": defaultdict(int)}
        
        for obj, (rel_type, dist) in relation['nearby_objs'].items():
            obj_cate = obj.category
            if not rel_type in relation_sets:
                relation_sets[rel_type] = defaultdict(int)
            
            relation_sets[rel_type][obj_cate] += 1

        for rel_type, item in relation_sets.items():
            relations[rel_type].append(item)
            if obj_id == target:
                current_relation[rel_type] = item

    # category
    if len(categories) > 1:
        return "category", categories
    
    # rooms
    if len(rooms) > 1:
        return "room", rooms
    
    # relations
    for rel_type, obj_list in relations.items():
        n = len(candidates) - len(obj_list)
        for i in range(n):
            obj_list.append(defaultdict(int))
        
        if not rel_type in current_relation:
            return "relation", (rel_type, "nothing")
        
        obj_dict_current = current_relation[rel_type]
        
        for i, obj_dict_a in enumerate(obj_list):
            for cate in obj_dict_current.keys():
                if cate not in obj_dict_a: # one of the
                    return "relation", (rel_type, cate)  
    
    # appearance
    return "appearance", None
\end{minted}
\end{algorithm}
% }
% \end{center}

% \begin{center}
%     \scalebox{0.75}{
    \begin{algorithm}
    \caption{This function returns the request information of the target object.}
    \label{alg:info}
    \begin{minted}[fontsize=\tiny]{python}
def get_info(self, object_id: str, info: tuple[str, list]):
    """
        return: info dict[str, any]
    """
    item_rel = self.spatial_relations[object_id] # scene graph
    item_attr = self.attribute_set[object_id] # object annotations
    info_type, info_content = info
    if info_type == 'category':
        return { "cate": item_rel["category"] }
    
    if info_type == 'room':
        return { "room": item_rel["room"] }
    
    if info_type == 'relation':
        rel_type_target, target_cate = info_content
        relations = item_rel["nearby_objects"]
        flag = False
        rel_type_set = set()
        for obj, (rel_type, dist) in relations.items():
            rel_type_set.add(rel_type)
            if rel_type_target == rel_type and target_cate == obj.category:
                flag = True
                break

        if target_cate == 'nothing':
            if rel_type_target in rel_type_set:
                return {"relation": [(False, rel_type_target, "nothing")]}
            else:
                return {"relation": [(True, rel_type_target, "nothing")]}

        return {"relation": [(flag, rel_type_target, target_cate)]}

    len_item_attr = len(item_attr)
    while len(self.sampled) < len_item_attr:
        attr_idx = np.random.randint(len_item_attr)
        _, attr = item_attr[attr_idx]
        if not attr in self.sampled:
            self.sampled.add(attr)
            break
    
    return {"appearance": [attr]}
\end{minted}
\end{algorithm}
% }
% \end{center}

% \begin{center}
%     \scalebox{0.75}{
\begin{algorithm}
    \caption{This function returns the subset that meets the condition given the original object set.}
    \label{alg:filter}
    \begin{minted}[fontsize=\tiny]{python}
def filter(self, candidates: set[str], condition: dict[str, list]):
    '''
        return: res_candidates set[str]
    '''
    res_candidates = set()
    candidates = [(obj_id, self.spatial_relations[obj_id]) # scene graph
                                                for obj_id in candidates]
    for obj_id, relation in candidates:
        attrs = self.attribute_set[obj_id] # object annotations
        if 'category' in condition:
            cate_condition = condition['category']
            cate_obj = relation['category']
            if not cate_condition == cate_obj:
                continue
        if 'room' in condition:
            room_condition = condition['room']
            room_obj = relation['room']
            if not room_condition == room_obj:
                continue
        if 'relation' in condition:
            relation_info = condition['relation']
            relation_item = relation['nearby_objects']
            flag = True
            for rel_info in relation_info:
                if flag == False:
                    break
                has_or_not, rel_a, cate_a = rel_info
                flag_has = False
                rel_b_set = set()
                for cate_b, (rel_b, dist) in relation_item.items():
                    rel_b_set.add(rel_b)
                    if rel_a == rel_b and cate_a == cate_b:
                        flag_has = True
                        flag = False
                        break
                if has_or_not:
                    if (cate_a == 'nothing' and rel_a in rel_b_set) or flag_has == False: 
                        flag = False 
                else:
                    if cate_a == 'nothing': 
                        if not rel_a in rel_b_set:
                            flag = False
            if not flag: # does not meet the condition
                continue
        if 'appearance' in condition:
            app_info = condition['appearance']
            app_item = attrs
            flag = False
            eb_a = self.get_embedding(app_info)
            for score, app_b in app_item:
                eb_b = self.get_embedding[app_b]
                if calc_similarity(eb_a, eb_b) > 0.9:
                    flag = True
                    break
            if not flag:
                continue

        res_candidates.add(obj_id)
    
    return res_candidates
\end{minted}
\end{algorithm}
% }
% \end{center}

\section{More Details of GRBench}
In this section, we provide more details about our three benchmarks in terms of episode generation (Sec.~\ref{sec:episode generation}), instruction generation (Sec.~\ref{sec:instruction generation}), and additional setup details for GRBench, including metrics implementation and simulation setups (Sec.~\ref{sec: more details benchmark setups}).

\subsection{Episode Generation}\label{sec:episode generation}
\begin{figure*}
\vspace{-1ex}
     \centering
          \begin{subfigure}[b]{0.32\textwidth}
         \centering
         \includegraphics[width=\textwidth]{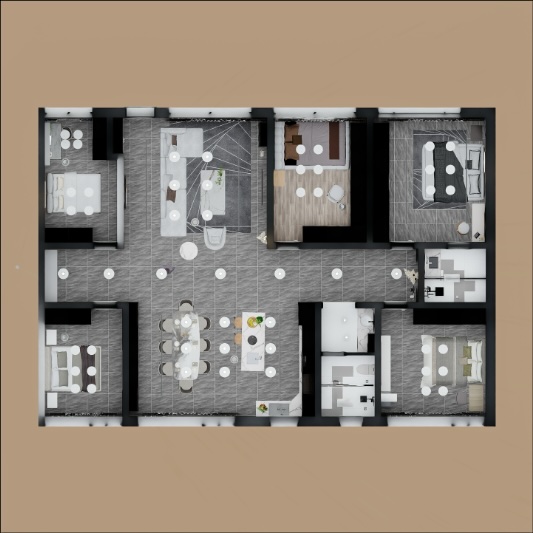}
         \caption{Bird's-Eye View (BEV) map.}
         \label{fig:topdown} 
     \end{subfigure}
     \hfill
     % \begin{subfigure}[b]{0.64\textwidth}
     %     \centering
     %     \includegraphics[width=\textwidth]{figures/interactive_objs.jpg}
     %     \caption{Interactive Objects Statistics with Part Annotations.}
     %     \label{fig:instance-stat} 
     % \end{subfigure}
     % \\[0.05in]
     % \begin{subfigure}[b]{0.34\textwidth}
     %    \centering
     %    \includegraphics[width=\textwidth]{figures/curated_scenes.jpg}
     %    \caption{Regions Statistics.}
     %    \label{fig:3d-box-prompt-stat} 
     % \end{subfigure}
     \begin{subfigure}[b]{0.32\textwidth}
         \centering
         \includegraphics[width=\textwidth]{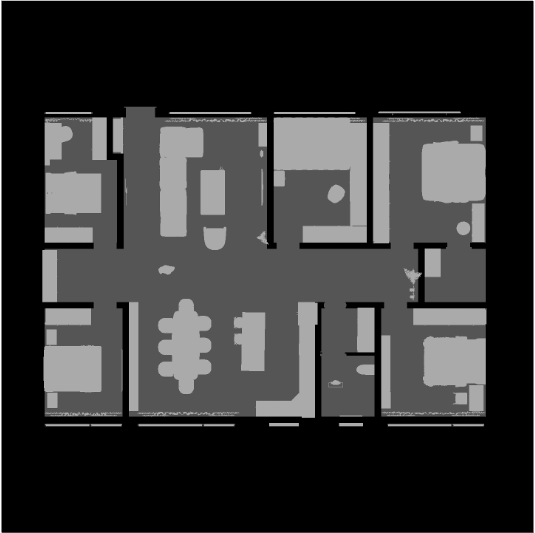}
         \caption{Occupancy map.}
         \label{fig:occupancy} 
     \end{subfigure}
     \hfill
      \begin{subfigure}[b]{0.32\textwidth}
         \centering
         \includegraphics[width=\textwidth]{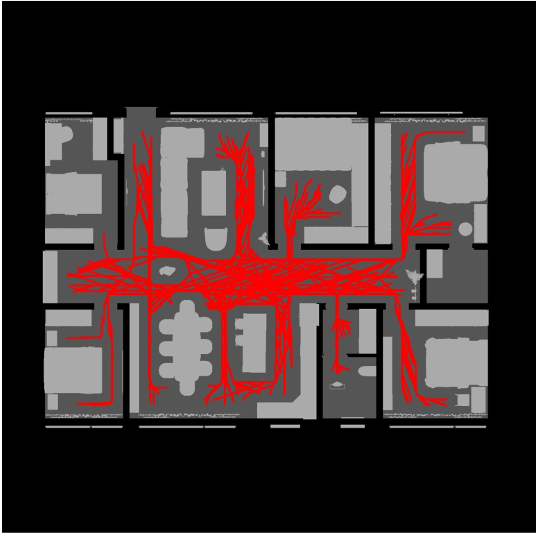}
         \caption{Collision-free paths.}
         \label{fig:paths} 
     \end{subfigure}
     \caption{Two steps of generating collision-free paths: 1) Generating the occupancy map of the scene. (The black/dark gray/light gray grids indicate the undefined/passable/obstacle regions respectively.) 2) Sampled collision-free paths on the occupancy map.} 
     \label{fig:path-sampling}
     % \vspace{-5ex}
\end{figure*}

\noindent\textbf{Sampling Valid Targets.} The starting points of the agent and the target object in the navigation episode are sampled, requiring the guarantee of solvability, i.e., a collision-free path must exist from the starting location to the target location. To achieve this, we first generate the occupancy map (Fig.~\ref{fig:occupancy}) for each scene by projecting all objects with a height between $[0.1,2.1]$ m in the scene onto the ground plane as occupied grids; grids outside the floor are marked as an undefined region, both of which are impassable. The remaining non-occupied grids represent passable areas. The resolution of the occupancy map is $1440\times1440$, with each pixel representing a unit length of $1.4$ cm. Using the occupancy map, we then generate collision-free paths (Fig.~\ref{fig:paths}) from randomly sampled points to the locations of objects. The radius for collision detection is $34$ cm\footnote{The radius of Unitree H1 is about $30.5$ cm. \url{https://www.unitree.com/h1/}}. The sampled paths should have a length between $[7,20]$ m to maintain a moderate task horizon length. These paths are used as ground-truth paths for navigation episodes, and object-centric task specifications are generated with our NPC. For objects that do not have a path meeting the criteria, we exclude them from the current version of our benchmarks.

\noindent\textbf{Sampling Conditions for Loco-Manipulation.} For Loco-Manipulation, we generate the ``pick-and-place'' episodes by sampling the handheld object and the target placement conditions. The source handheld object is sampled as the valid target sampling. The target placement locations are defined as conditions (target object and one spatial relation). We first sample the placement conditions from four potential spatial relations from $\left\{ on, nearby, nearby\times nearby, on\times nearby \right\}$. Notably, for each condition, we randomly sample the target objects from all candidates. If there are two conditions, we ensure the sampled two nearby objects (within 1.5 meters). For the ``on'' condition, we ensure the sampled object belongs to receptacle types. To simulate the daily life situation of multiple solutions, during instruction generation, we randomly drop some attributes of the target object description (``room'' or ``relation''). For evaluation, the satisfaction of conditions is assessed using the same spatial relation program as the generation process.

\subsection{Instruction Generation}\label{sec:instruction generation}
Given the generated episodes for three benchmarks, we detail the instruction generation process by GPT-4 in this section. The pseudo code for generating instructions is shown in Fig.~\ref{fig:enter-label}.

\noindent\textbf{Object Loco-Naivgation.} In Object Loco-Navigation, the goal is specified by a language instruction that describes the target object. To generate an instruction that can exclusively describe the target object, we need to find a set of attributes that can identify the target object without ambiguity. We achieve this goal by calling \texttt{\funcc{find\_diff}} and \texttt{\funcc{filter}} iteratively until the candidate set has only the target object, and collecting the searched information during the loop\footnote{For the sake of high generation quality and stability, we replaced the LLM-based programmer of our NPC with a static procedure to extract world knowledge. We manually programmed different procedures for the generation of different tasks.}. The collected information is then fed to the LLM-based speaker to generate the unique description of the target object. Due to the priority of difference searching, the generated instruction is concise and consistent with human speech habits.

\noindent\textbf{Social Loco-Navigation.} In Social Loco-Navigation, the given language instruction about the target object is often coarse and ambiguous, which is common in human daily dialogues. We aim to reproduce this uncertainty and thus require the agent to actively interact with the NPC efficiently to gather more concrete clues about the target. Since the difference searching has a coarse-to-fine priority, we can generate a coarse instruction by simply calling \texttt{\funcc{find\_diff}} once to obtain a crucial attribute that makes the target more distinct. We also have a probability to accept the last attribute searched through multi-round \texttt{\funcc{find\_diff}} and \texttt{\funcc{filter}} calling to enhance the diversity of instructions. This attribute is then sent to the speaker to generate the instruction.
\begin{figure}
    \centering
    \begin{subfigure}[t]{0.45\textwidth}
        \begin{minted}[fontsize=\tiny]{python}
# Object Loco-Navigation
def obj_nav(self, cands, target):
    infos = []
    while len(cands)>1:
        diff = self.find_diff(target, cands)
        info = self.get_info(target, diff)
        cands = self.filter(cands, info)
        infos.append(info)
    instruction = self.speaker(infos)
    return instruction

# Social Loco-Navigation
def social_nav(self, cands, t):
    rounds = np.random.randint(1, n)
    cnt = 0
    while len(cands) > 1 and cnt < rounds:
        diff = self.find_diff(t, cands)
        info = self.get_info(t, diff)
        cands = self.filter(cands, info)
        cnt += 1
    instruction = self.speaker([info])
    return instruction
    \end{minted}
    \end{subfigure}
    \hfill
    \begin{subfigure}[t]{0.50\textwidth}
        \begin{minted}[fontsize=\tiny]{python}
#  Loco-Manipulation
def loco_mani(self, cands, src, target_and_conds):
    """target_and_conds: a list of \\
        (possible_target, conditions)
    """
    # obtain the instruction for handheld object
    nav_instruction = self.obj_nav(cands, src)
    # obtain the instruction for target receptacle
    target_and_conds = target_and_conds.copy()
    all_infos = []
    for target, cond in target_and_conds:
        infos = []
        temp_cands = cands.copy()
        while len(temp_cands)>1:
            diff = self.find_diff(target, temp_cands)
            info = self.get_info(target, diff)
            temp_cands = self.filter(temp_cands, info)
            # randomly drop room or relation attributes
            if 'room' in info or 'relation' in info:
                if np.random.rand() > 0.5:
                    infos.append(info)
        all_infos.append((infos, cond))
    instruction = self.speaker(all_infos, nav_instruction)
    return instruction

    \end{minted}
    \end{subfigure}
    \caption{Pseudo code of generating instructions for three tasks.}
    \label{fig:enter-label}
\end{figure}

% Dialog Generation

\noindent\textbf{Loco-Manipulation.} In Loco-Manipulation, the goal is specified by several conditions, with each task defined by up to two placement conditions describing the target location of handheld objects. Each condition is paired with a specified spatial relation to target objects. The instruction generation process for these target locations, given the sampled conditions, proceeds as follows: 1)	The instruction for initial handheld object navigation is generated as Object Loco-Navigation. 2) For each placement condition, information for each sampled target object is collected by calling \texttt{\funcc{find\_diff}} and \texttt{\funcc{filter}}, following the prior benchmark. 3) For each collected piece of information, some attributes, including “room” and “relation,” are randomly dropped to simulate multiple solutions while ensuring the existence of a solution through the presence of target objects. The collected information, paired with the sampled relation condition and the initial handheld object navigation, is then fed into the LLM-based speaker to generate the final instruction.

\subsection{More Details of Benchmark Setups}\label{sec: more details benchmark setups}

\noindent\textbf{Metrics.} In addition to commonly used navigation metrics such as PL, SR, and SPL, we design two novel metrics (Sec.\red{4.2}). \textbf{ECR} assesses the efficiency that the dialogue helps to alleviate ambiguous candidates in Social Loco-Navigation:
\begin{equation}
\begin{aligned}
    ECR = \frac{\sum_{i=1}^n |\text{objects}_{i-1}|-|\text{objects}_i|}{|\text{objects}_0| - 1}, \\
    \text{objects}_i = \text{filter}(\text{objects}_{i-1}\text{, condition}_{i}),
\end{aligned}
\end{equation}
where $n$ is the number of dialogue rounds, condition$_i$ is the new constraint obtained from the $i$-th round dialogue, objects$_i$ is the subset of objects$_{i-1}$ that meets condition$_i$, and objects$_0$ is the set of all objects belonging to the category of the target object. \textbf{SCR} evaluates the fine-grained task progress of an episode in the Loco-Manipulation task:
\begin{equation}
    SCR = \frac{\sum_{i=1}^{n}\bm{1}(\text{condition}_i)}{n},
\end{equation}
where $n$ is the number of conditions, $\text{condition}_i$ indicates whether the $i$-th condition is satisfied, $\bm{1}(\cdot)$ is the indicator function that returns 1 when the input condition is satisfied, otherwise it returns 0.

\noindent\textbf{Simulation Setups.} The dt of the physical simulation is $1/240$ second, which is aligned with the training setting of control policies. Since the rendering process is the main bottleneck of simulation efficiency, we adopt different working frequencies for the low-level controller and the high-level planner. The dt of rendering is assigned as $1$ second, which means the agent can perform vision-dependent high-level planning at a frequency of 1 Hz. The control policy works at a frequency of 60 Hz.

\section{Baseline Implementation}

\subsection{Zero-Shot VLM Baseline}
In this section, we demonstrate the prompts we used for building zero-shot VLM agents.

\begin{tcolorbox}[title={Prompt Detail of the Object Loco-Navigation VLM Baseline}]
\begin{Verbatim}[breaklines=true,commandchars=\\\{\}]

\textcolor{purple}{You are a robot to find an object in the environment given the question:} \textcolor{blue}{\{question\}}

\textcolor{purple}{Your task is to select the most appropriate action based on the robot's current view. Here is the action list:}
1. Move forward 2 meters
2. Move forward 4 meters
3. Move forward 5 meters
4. Advance 2 meters to the left
5. Advance 4 meters to the left
6. Advance 6 meters to the left
7. Advance 2 meters to the right
8. Advance 4 meters to the right    
9. Advance 6 meters to the right
10. Turn left 90 degrees
11. Turn right 90 degrees
12. Stop

\textcolor{purple}{You must follow the following action selection conditions:}
1-9: If you think this action will help you to have a better view or get closer to the target object.
10-11: Only choose these two actions if it's necessary.
12:  If you think you find the target object and you are close enough to the target object.

\textcolor{purple}{You must follow the following rules:}
1. Select the action by only output the number with the action. For example: If you want to choose "Stop", just output "12".
2. Your output must only be a single integer from 1 to 12.
3. Never explain your choice.
4. Never include information in your answer that is not relevant to the question.
5. Robot's current view is the given image.
6. You can't continuously choose to turn right or turn left for more than 2 times. Now you have continuously chosen turn right or turn left for \textcolor{blue}{\{turning_time\}} times.
\end{Verbatim}
\end{tcolorbox}
\clearpage

\begin{tcolorbox}[title={Prompt Detail of the Social Loco-Navigation VLM Baseline}]
\begin{Verbatim}[breaklines=true,commandchars=\\\{\}]
\textcolor{purple}{Task Introduction:} You are a robot tasked with finding an object in an environment. Your task is to select the most appropriate action from 13 possible actions based on the robot's current view and known information about the target object in order to complete the task of finding the target object.

\textcolor{purple}{Action List:}
1. Move forward 2 meter
2. Move forward 4 meters
3. Move forward 5 meters
4. Advance 2 meter to the left
5. Advance 4 meters to the left
6. Advance 6 meters to the left
7. Advance 2 meter to the right
8. Advance 4 meters to the right    
9. Advance 6 meters to the right
10. Turn left 90 degrees
11. Turn right 90 degrees
12. Stop
13. Ask

\textcolor{purple}{Action Selection Conditions:}
1-9: If you think this action will help you to have a better view or get closer to the target object.
10-11: Only choose these two actions if it's necessary.
12:  If you think you find the target object and the target object is in the center of your field of view and you are close enough to the target object.
13: If you want more information about the target object.

\textcolor{purple}{Task:} \textcolor{blue}{\{task\}}

\textcolor{purple}{Target Object Information:} \textcolor{blue}{\{goal_info\}}

\textcolor{purple}{Request for Action Selection:} Based on the above information and robot's current view, please select the most appropriate action.

\textcolor{purple}{Follow these rules:}
1. Select the action by only output the number with the action.(Example: If you want to choose Stop, just output '13')
2. If your choice is Ask, you should also output the question you want to ask. Ensure to use a colon as the delimiter.(Example output: 13:Could you please tell me more information about the goal object?)
3. If your choice is not Ask, your output must only be a single integer from 1 to 12.
4. Never explain your choice.
5. Never include information in your answer that is not relevant to the question.
6. Robot's current view is the given image.
7. You can't continuously choose to turn right or turn left for more than 2 times. Now you have continuously chosen turn right or turn left for \textcolor{blue}{\{turning_time\}} times.
\end{Verbatim}
\end{tcolorbox}

\subsection{LLM Agent Baseline}

As shown in Fig.~\ref{fig:agent_overview}, the proposed LLM agent consists of a grounding module, a memory module, a decision module, and an action module. The environmental inputs to the agent are egocentric observations of the agent and the current state of the robot. Through the collaborative interactions among these modules, the agent can effectively analyze and utilize environmental inputs, enabling it to engage in both physical and linguistic interactions with the environment.

\noindent\textbf{Grounding Module} is responsible for processing raw environmental inputs into semantically rich information. It takes egocentric RGB-D images captured by the agent and the robot state as inputs and outputs corresponding semantic segmentation results, 3D point cloud, and 2\&3D bounding boxes for candidates. We employ SegFormer \cite{xie2021segformer} to perform initial segmentation on the RGB images. Then, combined with the segmentation result, we can use RGB-D images and the robot state to compute the point cloud in the current view and the point cloud for candidates if they exist. By bounding the point cloud of candidates, we can obtain 3D bounding boxes and, with projection to the 2D image plane, 2D bounding boxes can also be obtained.

\noindent\textbf{Memory Module} is responsible for maintaining the BEV map, action-observation history, and information about the target object obtained from dialogues between the agent and NPC. It provides directly usable information for decision and action modules. The BEV map is a 2D occupancy map containing candidate positions and descriptions, where the candidate positions are provided by the grounding module and descriptions are generated by a large model. The 2D occupancy map is updated in real-time using the 3D point cloud produced by the grounding module.

\noindent\textbf{Decision Module} is responsible for selecting the next action of the robot based on the information provided by the memory module. This function is primarily realized through a large model. This module has two main abilities: reasoning and speaking. When reasoning, the large model uses the prompt illustrated in \ref{prompt:reasoning} to choose the next navigation goal from the current candidates. If speaking, the model is in charge of generating a question to ask the NPC, as illustrated in the prompt in \ref{prompt:speaking}.

\noindent\textbf{Action Module} consists of 2 capabilities, navigation and manipulation. 
1) The navigation part makes real-time planning of a navigable path to the chosen navigation target. This is primarily achieved by using the RRT* algorithm. Since the 2D occupancy map used for path planning is updated in real-time, the navigation part replans a new path whenever the original path collides with the updated map, continuing until the robot successfully reaches the target or no viable path remains. 
2) The manipulation part executes actions like pick and place. Given the target 3D position, the manipulator uses an inverse-kinematics (IK) solver to plan the motion trajectory in joint space. To ease the difficulties of physically realistic picking, the target object can be directly attached to the gripper when the distance between them is within a certain threshold.

\begin{tcolorbox}[title={D.2: Prompt Detail of Reasoning in Decision Module}, label={prompt:reasoning}]
\begin{Verbatim}[breaklines=true,commandchars=\\\{\}]
\textcolor{purple}{USER:}
Here are the descriptions of the current candidates for the goal object \textcolor{blue}{\{goal\}}:

\textcolor{blue}{\{description\}}

Here are the known information about the goal object \textcolor{blue}{\{goal\}}:

\textcolor{blue}{\{goal_info\}}

1. Each line of candidate description corresponds to a candidate. 
2. The number in the description is the candidate's index, and the text after ':' is the candidate's description.
3. Now, based on the provided information about the goal object, please select the candidate most likely to be the goal object.
4. You only need to output the candidate's index. Please do not output anything other than the candidate's index.
\textcolor{purple}{ASSISTANT:}
\end{Verbatim}
\end{tcolorbox}

\begin{tcolorbox}[title={D.2: Prompt Detail of Speaking in Decision Module}, label={prompt:speaking}]
\begin{Verbatim}[breaklines=true,commandchars=\\\{\}]
\textcolor{purple}{USER:}
Here are the descriptions of the current candidates for the goal object \textcolor{blue}{\{goal\}}:

\textcolor{blue}{\{description\}}

Here are the known information about the goal object \textcolor{blue}{\{goal\}}:

\textcolor{blue}{\{goal_info\}}

1. Now you can ask a question about the goal object.
2. Based on the information described above,  what question do you think will help to minimize the scope of the possible candidates?
3. Just output the question, don't include the reason or explanation.
\textcolor{purple}{ASSISTANT:}
\end{Verbatim}
\end{tcolorbox}

\section{Real-world Demo}

We also demonstrate our LLM agent baseline on Object Loco-Navigation in the real world. In the demo, the locomotion capability of H1 is powered by the same control policy used in the simulation. It shows that the agents designed in our simulation platform can be smoothly transferred to a real-world robot driven by our sim-to-real-capable controller. We will release the real-world demo after further polishing in the near future.

\end{document}